%% file: cameral_ready.tex
\documentclass[nohyperref]{article}

\usepackage{microtype}
\usepackage{graphicx}
\usepackage{subfigure}
\usepackage{booktabs} 

\usepackage{amsfonts}
\usepackage{capt-of}
\usepackage{enumitem}
\usepackage{makecell}
\newcommand*\tcircle[1]{%
  \raisebox{-0.5pt}{%
    \textcircled{\fontsize{7pt}{0}\fontfamily{phv}\selectfont #1}%
  }%
}

\usepackage{hyperref}

\usepackage[accepted]{icml2022}

\usepackage{amsmath}
\usepackage{amssymb}
\usepackage{mathtools}
\usepackage{amsthm}

\usepackage[capitalize,noabbrev]{cleveref}

\theoremstyle{plain}
\newtheorem{theorem}{Theorem}[section]

\newtheorem{lemma}[theorem]{Lemma}

\theoremstyle{definition}

\newtheorem{assumption}[theorem]{Assumption}
\theoremstyle{remark}
\newtheorem{remark}[theorem]{Remark}

\usepackage[textsize=tiny]{todonotes}

\icmltitlerunning{Detached Error Feedback for Distributed SGD with Random Sparsification}

\begin{document}

\twocolumn[
\icmltitle{Detached Error Feedback for Distributed SGD with Random Sparsification}



\icmlsetsymbol{equal}{*}

\begin{icmlauthorlist}
\icmlauthor{An Xu}{yyy}
\icmlauthor{Heng Huang}{yyy}
\end{icmlauthorlist}

\icmlaffiliation{yyy}{Department of Electrical and Computer Engineering, University of Pittsburgh, Pittsburgh, PA 15213, USA. This work was partially supported by  NSF IIS 1845666, 1852606, 1838627, 1837956, 1956002, IIA 2040588}

\icmlcorrespondingauthor{Heng Huang}{heng.huang@pitt.edu}


\icmlkeywords{Machine Learning, ICML}

\vskip 0.3in
]



\printAffiliationsAndNotice{}  

\begin{abstract}
The communication bottleneck has been a critical problem in large-scale distributed deep learning. In this work, we study distributed SGD with random block-wise sparsification as the gradient compressor, which is ring-allreduce compatible and highly computation-efficient but leads to inferior performance. To tackle this important issue, we improve the communication-efficient distributed SGD from a novel aspect, that is, the trade-off between the variance and second moment of the gradient. With this motivation, we propose a new detached error feedback (DEF) algorithm, which shows better convergence bound than error feedback for non-convex problems. We also propose DEF-A to accelerate the generalization of DEF at the early stages of the training, which shows better generalization bounds than DEF. Furthermore, we establish the connection between communication-efficient distributed SGD and SGD with iterate averaging (SGD-IA) for the first time. Extensive deep learning experiments show significant empirical improvement of the proposed methods under various settings.
\end{abstract}

\section{Introduction}\label{introduction}

Deep learning models are hard to train due to the heavy computation complexity and long training iterations. Therefore, distributed deep learning with multiple workers (GPUs) has become a prevalent practice to parallelize and accelerate the training for large-scale tasks, where the model and dataset sizes continue to grow nowadays \cite{simonyan2014very,he2016deep,deng2009imagenet}.

Nevertheless, synchronous distributed training have difficulty in scaling up the number of workers for large deep learning models, as the gradient in each worker to be communicated per iteration is of the same dimension as the model size. It is also known as the communication bottleneck. Besides, it incurs imbalanced communication traffic in the parameter-server \cite{li2014scaling,li2014communication,li2013parameter} architecture, where the server suffers from much larger communication burden than workers. To address the communication bottleneck issue, there have been numerous lines of works including asynchronous execution \cite{dean2012large}, gradient compression \cite{bernstein2018signsgd,bernstein2018signsgd2,wen2017terngrad,alistarh2017qsgd,aji2017sparse,alistarh2018convergence,strom2015scalable,lin2018deep,gao2021convergence}, communication scheduling \cite{george2020distributed}, infrequent communication \cite{stich2018local}, delayed gradient \cite{NEURIPS2018_2c6a0bae,zhu2021delayed}, decentralized training \cite{lian2017can,koloskova2019decentralized,tang2018d,assran2019stochastic,koloskova2019decentralized}, model parallelism \cite{huang2019gpipe,xu2020acceleration}, etc.

In this work, we focus on synchronous distributed SGD with gradient compression, or more specifically, random block-wise gradient sparsification (RBGS) \cite{vogels2019powersgd,xie2020cser}. The most popular gradient sparsifier is probably the Top-$\mathcal{K}$ gradient sparsification \cite{alistarh2018convergence,lin2018deep}, where each worker selects the largest $\mathcal{K}$ gradient components according to the absolute value as the sparsified gradient. However, Top-$\mathcal{K}$ has several drawbacks: 1) it requires extra communication overheads to communicate the gradient indices, 2) it is applied in parameter-server architecture but not ring-allreduce compatible, and most of all, 3) its computation overheads $\mathcal{O}(\mathcal{K}\log_2 d)$ for model $\theta\in\mathbb{R}^d$ may even outweigh its communication benefits \cite{song2021communication,xie2020cser,sahu2021rethinking} as it is efficient only for a small $\mathcal{K}$ for optimized implementations on GPU \cite{shanbhag2018efficient}. While in RBGS, we randomly sample a block of gradient as the sparsified gradient for communication among workers. To ensure the consistency of the sampling process, each worker will be pre-assigned the same random seed. In comparison to Top-$\mathcal{K}$, RBGS is highly computation-efficient ($\mathcal{O}(1)$) as we only need to uniformly and randomly sample one starting index of the gradient block. RBGS is also ring-allreduce compatible. However, RBGS results in inferior model performance in that its sparsified gradient usually does not include as many significant gradient components as Top-$\mathcal{K}$, leading to large compression error.

To address this important problem, we propose a novel detached error feedback method (DEF), while the vanilla error feedback (EF) method \cite{karimireddy2019error,zheng2019communication} fails to address it. We summarize our major contributions as follows.
\begin{itemize}[leftmargin=0.12in]
\setlength{\itemsep}{0pt}
    \item Our proposed DEF method is motivated by a novel insight that a trade-off between the gradient variance and second moment can improve the convergence bound related to compression error.
    \item We propose DEF-A to accelerate the generalization during the training with support from corresponding generalization analysis. It potentially demystifies why compression helps to improve the performance in some prior works \cite{avdiukhin2021escaping,zhao2020feature,bernstein2018signsgd,bernstein2018signsgd2}.
    \item We find that SGD with iterate averaging (SGD-IA) \cite{polyak1990new,ruppert1988efficient,neu2018iterate,wu2020obtaining} can be viewed as a special case of communication-efficient distributed SGD for the first time. Consequently, our generalization analysis of DEF-A extends to SGD-IA, providing potential theoretical explanations for some other applications incorporating SGD-IA \cite{he2020momentum,izmailov2018averaging,huang2017snapshot}.
    \item Extensive deep image classification experiments on CIFAR-10/100 and ImageNet show significant improvements of DEF(-A) over existing works with RBGS.
\end{itemize}

\section{Related Works}
To begin with, suppose the training dataset $\mathcal{S}=\{\xi_n\}_{n=1}^{N}$ and we have the training objective function
\begin{equation}
    F_{\mathcal{S}}(\theta)=\frac{1}{N}\sum^{N}_{n=1}f(\theta;\xi_n)=\mathbb{E}_{\xi\in\mathcal{S}}f(\theta;\xi)
\end{equation}
to minimize, where $\theta\in\mathbb{R}^d$ denotes the model and $f$ is the loss function. From now on, we will omit the subscript in $\mathbb{E}$ if the context is clear. For distributed SGD at iteration $t$, each worker $k$ randomly selects one data sample $\xi_{k,t}\in\mathcal{S}$ and computes the stochastic gradient $g_{k,t}=\nabla f(\theta_t;\xi_{k,t})$. Then all the workers communication to get the average gradient $g_t=\frac{1}{K}\sum^{K}_{k=1}g_{k,t}$, where $K$ is the total number of workers, and update the model via
\begin{equation}
    \theta_{t+1}=\theta_t-\eta g_t \,,
\end{equation}
where $\eta$ is the learning rate.

\textbf{Compression.} Gradient compression includes quantization \cite{bernstein2018signsgd,bernstein2018signsgd2,wen2017terngrad,alistarh2017qsgd}, which reduces the 32-bit gradient component to as low as 1 bit (compression ratio $\leq 32$), and sparsification \cite{aji2017sparse,alistarh2018convergence,strom2015scalable}, which reduces the number of gradient components for communication. Let the compression function be $\mathcal{C}$, then the workers will communicate $\mathcal{C}(g_{k,t})$ instead of $g_{k,t}$. In general, sparsification achieves flexible and higher compression ratio than quantization. Besides Top-$\mathcal{K}$, random-$\mathcal{K}$ \cite{elibol2019variance,stich2018sparsified} randomly selects $\mathcal{K}$ gradient components as the sparsified gradient. \citet{dutta2020discrepancy} selects gradient components larger than a threshold and is a variable-dimension compressor. \citet{wangni2018gradient,song2021communication} propose to select each gradient component with a probability to keep the sparsified gradient unbiased. In this work, we consider RBGS \cite{vogels2019powersgd,xie2020cser}, which is most easy to implement, highly computation-efficient, but challenging to retain the model performance. Moreover, it is ring-allreduce compatible for SOTA GPU communication backend library (e.g., NCCL), i.e.,
\begin{equation}
    \mathcal{C}(\Delta_1)+\mathcal{C}(\Delta_2)=\mathcal{C}(\Delta_1+\Delta_2) \,.
\end{equation}

\textbf{Error Feedback.} Error feedback (EF) \cite{karimireddy2019error,tang2019doublesqueeze} method maintains local compression error $e_{k,t}$ at worker $k$, adds it to the current gradient before compression, and communicates to average $\mathcal{C}(\eta g_{k,t}+e_{k,t})$. The error is updated via
\begin{equation}
    e_{k,t+1}=\eta g_{k,t}+e_{k,t}-\mathcal{C}(g_{k,t}+e_{k,t}) \,.
\end{equation}
\citet{zheng2019communication} extends EF to momentum SGD \cite{polyak1964some}. EF works well for Top-$\mathcal{K}$ sparsifier but poorly for RBGS. \citet{xie2020cser} proposes PSync to immediately apply local error to each worker's model for RBGS. However, we will show that PSync works better for Wide ResNet \cite{zagoruyko2016wide} but has scalability issue for other common model architectures. SAEF \cite{xu2021step} proposes to apply the local error before computing gradient in the next iteration to accelerate the generalization during training. Other EF variants includes EF21 \cite{richtarik2021ef21,fatkhullin2021ef21} which compresses the gradient difference \cite{mishchenko2019distributed} but is evaluated only on logistic regression problems, acceleration for EF \cite{qian2021error,li2020acceleration}, EF for variance reduction \cite{tang2021errorcompensatedx},  etc.

\textbf{Generalization Analysis.} The generalization analysis of this work incorporates the uniform stability \cite{bousquet2002stability,hardt2016train} approach, focusing on the inherent stability property of the learning algorithm. \citet{bousquet2002stability} analyzes bagging methods. It is later used to analyze the generalization property of SGD \cite{hardt2016train} and its momentum variants \cite{10.5555/3304889.3305071}. \citet{kuzborskij2018data} establishes a data-dependent notion of the stability to stress the distribution-dependent risk of the initialization point and make the generalization bounds more optimistic. \citet{zhou2021towards} analyzes the generalization of the Lookahead optimizer \cite{zhang2019lookahead} with uniform stability.

As there are numerous works combining various techniques \cite{basu2019qsparse}, in this work, we focus on random block-wise gradient sparsification (RBGS).

\section{Detached Error Feedback}

\begin{algorithm}[t]
    \caption{Detached Error Feedback (DEF(-A)).}
    \label{def algorithm}
\begin{algorithmic}[1]
    \STATE \textbf{Input:} training dataset $\mathcal{S}$, number of iterations $T$, number of workers $K$, learning rate $\eta$, ring-allreduce compressor $\mathcal{C}$,  coefficient $\lambda\in[0,1]$.
    \STATE \textbf{Initialize:} model $x_0=y_0$, local compression error $e_{k,0}=0$, worker $k\in[K]$.
    \FOR{$t=0,1,\cdots,T-1$}
        \FOR{worker $k\in[k]$ in parallel}
            \STATE Randomly sample data $\xi_{k,t}$ from $\mathcal{S}$.
            \STATE Compute $g_{k,t}=\nabla f(x_t-\lambda e_{k,t};\xi_{k,t})$. \hfill \textit{// detach}
            \STATE $p_{k,t}=\eta g_{k,t}+e_{k,t}$. \hfill \textit{// error feedback}
            \STATE $e_{k,t+1}=p_{k,t}-\mathcal{C}(p_{k,t})$.
            \STATE Ring-allreduce: $\mathcal{C}(p_t)=\mathcal{C}(\frac{1}{K}\sum^{K}_{k=1}p_{k,t})=\frac{1}{K}\sum^{K}_{k=1}\mathcal{C}(p_{k,t})$.
            \STATE Update $x_{t+1}=x_{t}-\mathcal{C}(p_t)$.
        \ENDFOR
    \ENDFOR
    \STATE \textbf{Output:} $y_T=x_T-e_T=x_T-\frac{1}{K}\sum^{K}_{k=1}e_{k,T}$ for DEF and $x_T$ for DEF-A.
\end{algorithmic}
\end{algorithm}

In this section, we described our proposed DEF method (Algorithm \ref{def algorithm}) in detail. As RBGS is a very aggressive compressor, the algorithm is crucial for better performance.

\subsection{Motivation}
In EF variants \cite{karimireddy2019error,zheng2019communication,xie2020cser} for practical large-scale distributed training of deep learning models, Assumptions \ref{bounded variance} and \ref{bounded second moment} are needed to bound the norm of the stochastic gradient
\begin{equation}
    \|\nabla f(\theta;\xi)\|_2\leq G=\sqrt{\sigma^2+M^2} \,.
\end{equation}
Then $G$ bounds the compression error
\begin{equation}
    \frac{1}{K}\sum^{K}_{k=1}\|e_{k,t}\|^2_2=\mathcal{O}(\sigma^2+M^2)
\end{equation}
at iteration $t$. Though Assumption \ref{bounded second moment} often appears in related literature, it is usually regarded as a strong assumption \cite{richtarik2021ef21} because $M^2$ could be much larger than $\sigma^2$. Hereby, we propose a novel insight that if some trade-off coefficient $\alpha$ can be introduced to transform the compression error bound to a similar interpolation form as
\begin{equation}
    (\alpha \sigma)^2 + ((1-\alpha)M)^2 \overset{\alpha=\frac{M^2}{\sigma^2+M^2}}{\geq} \frac{\sigma^2M^2}{\sigma^2+M^2}\overset{M\gg \sigma}{=} \sigma^2 \,,
\end{equation}
then the bound $\mathcal{O}(\sigma^2+M^2)$ can be reduced to $\mathcal{O}(\sigma^2)$ when $M\rightarrow\infty$, i.e., Assumption \ref{bounded second moment} does not hold.

\begin{assumption}\label{bounded variance}
(Bounded Variance) $\forall \theta\in\mathbb{R}^d$, the variance of the stochastic gradient satisfies $\mathbb{E}_{\xi\in\mathcal{S}}\|\nabla f(\theta; \xi)-\nabla F_{\mathcal{S}}(\theta)\|^2_2 \leq \sigma^2$.
\end{assumption}

\begin{assumption}\label{bounded second moment}
(Bounded Second Moment) $\forall \theta \in \mathbb{R}^d$, the second moment of the full gradient satisfies $\|\nabla F_{\mathcal{S}}(\theta)\|^2_2 \leq M^2$.
\end{assumption}

\subsection{Algorithm}

\begin{assumption}\label{allreduce compressor}
(Ring-allreduce Compressor) $\forall \Delta_1, \Delta_2 \in \mathbb{R}^d$, the compressor $\mathcal{C}$ satisfies $\mathcal{C}(\Delta_1)+\mathcal{C}(\Delta_2)=\mathcal{C}(\Delta_1 + \Delta_2)$.
\end{assumption}
Firstly, the ring-allreduce communication architecture requires that the compressor should satisfy Assumption \ref{allreduce compressor} such that Algorithm \ref{def algorithm} line 9 holds. RBGS satisfies this assumption.

Secondly, DEF returns $y_T=x_T-e_T$ by default because we have
\begin{equation}\label{eq:y update}
    y_{t+1}=y_t-\eta g_t=y_t-\frac{1}{K}\sum^{K}_{k=1}g_{k,t}\,.
\end{equation}
In particular, when $K=1$ (single worker) and $\lambda=1$, $\{y_t\}$ is identical to the SGD solution path. We note that averaging $e_T=\frac{1}{K}\sum^{K}_{k=1}e_{k,T}$ only incurs a one-time communication cost after the training concludes.

Then, a major difference of DEF and EF is that we evaluate gradient at $x_t-\lambda e_{k,t}$, a point \textbf{detached} from the point $x_t$ to evaluate gradient as in EF. This step does not incur any communication cost. From Eq.~(\ref{eq:y update}), our goal is to make sure that the point to evaluate gradient $g_{k,t}$ is as close to $y_t=x_t-e_t$ as possible. For EF, the distance is $\|x_t-y_t\|^2_2=\|e_t\|^2_2\leq \frac{1}{K}\sum^{K}_{k=1}\|e_{k,t}\|^2_2$, while for DEF, the average distance to minimize regarding $\lambda$ becomes
\begin{equation}\label{eq:min d}
    \frac{1}{K}\sum^{K}_{k=1}\|x_t-\lambda e_{k,t}-y_t\|^2_2=\frac{1}{K}\sum^{K}_{k=1}\|e_t-\lambda e_{k,t}\|^2_2 \,.
\end{equation}
(1) When $\lambda=\lambda(k,t)$, it is obvious that $\lambda^*(k,t)=\frac{\langle e_t, e_{k,t}\rangle}{\|e_{k,t}\|^2_2}$, which is determined by the \textit{projection} of $e_{t}$ onto $e_{k,t}$. However, it is impractical to decide $\lambda^*(k,t)$ for worker $k$ at iteration $t$ as $e_t$ is unknown ($e_t=\frac{1}{K}\sum^{K}_{k=1}e_{k,t}$ needs extra communication cost).

(2) When $\lambda=\lambda(t)$, we can derive $\lambda^*(t)=\frac{\|e_t\|^2_2}{\frac{1}{K}\sum^{K}_{k=1}\|e_{k,t}\|^2_2}$, which is still impractical due to unkown $e_t$.

(3) Therefore, we will regard $\lambda$ as a tuned hyper-parameter, invariant regarding $k$ and $t$. Then it becomes minimizing the sum of the errors $\frac{1}{KT}\sum^{T-1}_{t=0}\sum^{K}_{k=1}\|e_t-\lambda e_{k,t}\|^2_2$ which will appear in the convergence bound of DEF, similar to the suggestion in \citet{sahu2021rethinking}. Previously when $\lambda$ is a function of $t$, it reduces to minimizing Eq.~(\ref{eq:min d}). In our CIFAR-10 VGG-16 experiments with $\lambda=0.3$, we find that the new distance is $\times 1.7$ smaller than the distance in EF.

\textbf{Relation to Motivation.} Minimizing Eq.~(\ref{eq:min d}) is closely related to the motivation since
\begin{align}
    &\|e_t-\lambda e_{k,t}\|^2_2 = \|(\underbrace{\eta g_{t-1}-\lambda\eta g_{k,t-1}}+e_{t-1}-\lambda e_{k,t-1}) \notag\\
    &\quad - \mathcal{C}(\underbrace{\eta g_{t-1}-\lambda\eta g_{k,t-1}}+e_{t-1}-\lambda e_{k,t-1})\|^2_2 \,,
\end{align}
where $g_{t-1}-\lambda g_{k,t-1}$ is affected by the gradient variance and second moment trade-off via the choice of $\lambda$. For example, in extreme circumstances where $\sigma=0$, in expectation, local errors on different workers are the same and $g_{t-1}-\lambda g_{k,t-1}$ is zero with $\lambda=1$.


\textbf{Momentum Variant.} It is easy to extend DEF to momentum SGD variant. Let the momentum buffer on worker $k$ be $m_{k,0}=0$ and the momentum constant be $\mu$. We only need to substitute Algorithm \ref{def algorithm} line 7 with
\begin{equation}
    m_{k,t+1} = \mu m_{k,t} + g_{k,t},\, p_{k,t} = \eta m_{k,t+1} + e_{k,t} \,.
\end{equation}

\textbf{DEF-A.} Simply returning $x_T$ can accelerate the generalization performance of DEF during training in that when $K=1$, $\lambda=1$ and $\mathcal{C}(\Delta)=\delta\Delta$ ($0<\delta<1$), $\{y_t\}$ reduces to SGD and $\{x_t\}$ reduces to a special case of SGD-IA (Iterate Averaging, a combination of models in each iteration) \cite{wu2020obtaining}:
\begin{equation}\label{eq:special sgd-ia}
    x_t=\underbrace{(1-\delta)^t}_{P_0}y_0 + \sum^{t}_{t^\prime=1}\underbrace{\delta(1-\delta)^{t-t^\prime}}_{P_{t^\prime}}y_{t^\prime} \,,
\end{equation}
where $P_0+P_1+\cdots+P_t=1$. Note that for Polyak-Ruppert IA \cite{polyak1990new}, $P_0=P_1=\cdots=P_t=\frac{1}{t+1}$. While for geomeric Polyak-Ruppert IA \cite{neu2018iterate}, $P_{t^\prime}=\frac{\beta^{t^\prime}}{1+\beta+\cdots+\beta^t}$ where $0<\beta<1$ is some constant and $0\leq t^\prime\leq t$. However, this part is based on generalization analysis instead of convergence analysis as for DEF. Hence we leave the details of the general case in the next section.

\section{Theoretical Analysis}
In this section, we consider non-convex objective functions as our target is the deep learning model. All detailed proof can be found in the Appendix. Suppose that each $\xi_n$ in the training dataset $\mathcal{S}$ is i.i.d drawn from an unknown data distribution $\mathcal{D}$ and $F_{\mathcal{D}}(\theta)=\mathbb{E}_{\xi\in\mathcal{D}}f(\theta;\xi)$. For generalization, we are interested in how the model $\theta_{\mathcal{A},\mathcal{S}}$, which is trained on $\mathcal{S}$ with a randomized algorithm $\mathcal{A}$, generalizes on $\mathcal{D}$ by measuring the well-known excess risk error $\epsilon$.
\begin{equation}
\begin{split}
    \epsilon &= \mathbb{E}_{\mathcal{A},\mathcal{S}}[F_{\mathcal{D}}(\theta_{\mathcal{A},\mathcal{S}})] - \mathbb{E}_{\mathcal{A},\mathcal{S}}[F_{\mathcal{S}}(\theta^*_{\mathcal{S}})]\\
    &= \underbrace{\mathbb{E_{\mathcal{A},\mathcal{S}}}[F_{\mathcal{S}}(\mathcal{\theta}_{\mathcal{A},\mathcal{S}})-F_{\mathcal{S}}(\theta^*_{\mathcal{S}})]}_{\text{optimization error } \epsilon_{\text{opt}}} \\
    &\quad+ \underbrace{\mathbb{E}_{\mathcal{A},\mathcal{S}}[F_{\mathcal{D}}(\theta_{\mathcal{A},\mathcal{S}}) - F_{\mathcal{S}}(\theta_{\mathcal{A},\mathcal{S}})]}_{\text{generalization error } \epsilon_{\text{gen}}}
\end{split}
\end{equation}

\begin{assumption}\label{lipschitz gradient}
(L-Lipschitz Smooth) $\forall \theta_1, \theta_2 \in \mathbb{R}^d$, the loss function satisfies
\begin{equation}
    \|\nabla f(\theta_1;\xi)-\nabla f(\theta_2;\xi)\|_2\leq L\|\theta_1-\theta_2\|_2 \,.
\end{equation}
It also implies that
\begin{equation}
    \|\nabla F(\theta_1)-\nabla F(\theta_2)\|_2\leq L\|\theta_1-\theta_2\|_2 \,.
\end{equation}
\end{assumption}

\begin{assumption}\label{approximate compressor}
($\delta$-approximate Compressor) $\forall \Delta\in\mathbb{R}^d$, the compressor $\mathcal{C}$ satisfies \begin{equation}
    \|\mathcal{C}(\Delta)-\Delta\|^2_2 \leq (1-\delta)\|\Delta\|^2_2 \,,
\end{equation}
where $0<\delta<1$ is related to the compression ratio.
\end{assumption}
This assumption is widely used in communication-efficient distributed SGD \cite{karimireddy2019error,zheng2019communication,xie2020cser}. For RBGS, we can take an expectation over the random compression and $\delta$ will be identical to the compression ratio.

\subsection{Convergence Rate}
In this section, we bound the gradient norm $\|\nabla F(\theta_{\mathcal{A},\mathcal{S}})\|^2_2$ for convergence rate analysis of the proposed DEF method.

\begin{theorem}\label{def convergence}
(Convergence Rate of DEF, Appendix \ref{appendix:convergence}) Let Assumptions \ref{bounded variance}, \ref{bounded second moment}, \ref{allreduce compressor}, \ref{lipschitz gradient} and \ref{approximate compressor} hold. If $\eta\leq \frac{1}{4L}$, we have
\begin{align}\label{eq:def convergence}
    &\frac{1}{T}\sum^{T-1}_{t=0}\mathbb{E}\|\nabla F_{\mathcal{S}}(y_t)\|^2_2 \leq \frac{4\mathbb{E}[F_{\mathcal{S}}(y_0)-F_{\mathcal{S}}(y^*)]}{\eta T} + \frac{2\eta L \sigma^2}{K} \notag\\
    &+ \frac{4\eta^2L^2[\frac{K-1}{K^2}\sigma^2 + (\frac{1}{K}-\lambda)^2\sigma^2 + 2(1-\lambda)^2M^2]}{(\sqrt{(1-\delta/2)/(1-\delta)}-1)^2} \,.
\end{align}
\end{theorem}

\begin{remark}
Suppose $\theta_{\mathcal{A},\mathcal{S}}$ is randomly chosen from the sequence $\{y_t\}_{t=0}^{T-1}$, $\eta=\mathcal{O}(\sqrt{\frac{K}{T}})\leq \frac{1}{4L}$, and $K=\mathcal{O}(T^{1/3})$ (i.e., $T$ is large enough), we have $\mathbb{E}\|\nabla F_{\mathcal{S}}(\theta_{\mathcal{A},\mathcal{S}})]\|^2_2 = \mathcal{O}(\frac{1}{\sqrt{KT}} + \frac{K}{T})=\mathcal{O}(\frac{1}{\sqrt{KT}})$. It matches the rate of SGD with linear speedup regarding the number of workers $K$.
\end{remark}

\begin{remark}
The last term in Eq.~(\ref{eq:def convergence}) is determined by the compression error. When $\delta$, $\sigma$, $M$ are of interest, we have $\mathbb{E}\|\nabla F_{\mathcal{S}}(\theta_{\mathcal{A},\mathcal{S}})]\|^2_2=$
\begin{equation}\label{eq:compression error complexity}
    \mathcal{O}(\frac{\frac{K-1}{K^2}\sigma^2 + (\frac{1}{K}-\lambda)^2\sigma^2 + 2(1-\lambda)^2M^2}{(\sqrt{(1-\delta/2)/(1-\delta)}-1)^2}) \,.
\end{equation}
(1) When $K=1$ (single worker) and $\lambda=1$, it \textit{vanishes}, which is better than EF \cite{karimireddy2019error,zheng2019communication}.


(2) When $\sigma$ and $M$ are of interest, following the motivation in the previous section and ignoring other constant factors, Eq.~(\ref{eq:compression error complexity}) becomes
\begin{equation}
    \mathcal{O}(\frac{K-1}{K^2}\sigma^2 + \frac{2(1-\frac{1}{K})^2\sigma^2M^2}{\sigma^2 + 2M^2}) \,.
\end{equation}
when $\lambda=\frac{\frac{1}{K}\sigma^2 + 2M^2}{\sigma^2 + 2M^2}$. It further reduces to $\mathcal{O}(\frac{K-1}{K}\sigma^2)$ when $M\rightarrow \infty$ (i.e. Assumption \ref{bounded second moment} does not hold). Therefore, DEF is the first EF variant compressing gradient \textit{without relying on the bound of the gradient second moment}.

(3) When $K$ is large and $\sigma$ and $M$ are of interest, our bound improves $\mathcal{O}(\sigma^2+M^2)$ \cite{karimireddy2019error,zheng2019communication,xie2020cser} to
\begin{equation}
    \mathcal{O}(\frac{2\sigma^2M^2}{\sigma^2+2M^2}) \,.
\end{equation}
Our empirical deep learning experiments suggest that $\sigma^2\approx 0.3M^2$, which means that our bound is about $\times 5$ \textit{smaller} ignoring other constant factors.
\end{remark}

\subsection{Generalization Rate}
In this section, we consider non-convex objective functions under PL condition, which establishes the relation between the gradient norm and the optimization error $\epsilon_{\text{opt}}$ \cite{zhou2021towards}. We bound the excess risk error $\epsilon=\epsilon_{\text{opt}}+\epsilon_{gen}$ for the generalization analysis of the proposed DEF(-A) method.

\textbf{Polyak-\L{}ojasiewicz (PL) Condition} \cite{karimi2016linear}. Let $\theta^*\in\min_{\theta\in\mathbb{R}^d}F_{\mathcal{S}}(\theta)$. The objective function $F_{\mathcal{S}}(\theta)$ satisfies $\mu$-PL condition if $\forall \theta\in\mathbb{R}^d$, we have
\begin{equation}
    2\mu [F_{\mathcal{S}}(\theta)-F_{\mathcal{S}}(\theta^*)]\leq \|\nabla F_{\mathcal{S}}(\theta)\|^2_2 \,.
\end{equation}

\begin{theorem}\label{def generalization}
(Excess Risk Error of DEF(-A), Appendix \ref{appendix:gen}) Let Assumptions \ref{bounded variance}, \ref{bounded second moment}, \ref{allreduce compressor}, \ref{lipschitz gradient} and \ref{approximate compressor} hold. Suppose $\eta=\frac{c}{t+1}$, where $c>0$ is some constant.

(1) The generalization error of DEF
\begin{equation}\label{eq:def gen}
    \epsilon_{\text{gen}}=\mathcal{O}(T^{(1-\frac{K}{N})Lc/((1-\frac{K}{N})Lc+1)}) \,.
\end{equation}
(2) Suppose $\eta\leq\frac{1}{4L}$. The optimization error of DEF
\begin{equation}
    \epsilon_{\text{opt}} = \widetilde{\mathcal{O}}(T^{-\frac{\mu c}{2}}+T^{-1}) \,.
\end{equation}
(3) For RBGS, the generalization error of DEF-A
\begin{equation}\label{eq:def-a gen}
    \epsilon_{\text{gen}}=\mathcal{O}(T^{(1-\frac{K}{N})\delta^{\frac{1}{2}} Lc/((1-\frac{K}{N})\delta^{\frac{1}{2}} Lc+1)}) \,.
\end{equation}
(4) Suppose $\eta\leq\frac{1}{8L}$. The optimization error of DEF-A
\begin{equation}
    \epsilon_{\text{opt}} = \widetilde{\mathcal{O}}(T^{-\frac{\mu\delta c}{2}} + T^{-1} + (1/\sqrt{1-\delta}-1)^{-2}) \,.
\end{equation}
\end{theorem}
\begin{remark}
When $K=1$, Eq.~(\ref{eq:def gen}) matches the result of SGD in \citet{hardt2016train}.
\end{remark}
\begin{remark}
DEF-A has a better $\epsilon_{\text{gen}}$ but a worse $\epsilon_{\text{opt}}$ than DEF. Since $\epsilon=\epsilon_{\text{gen}}+\epsilon_{\text{opt}}$, DEF-A can achieve \textit{better} generalization rate than DEF via a trade-off between $\epsilon_{\text{gen}}$ and $\epsilon_{\text{opt}}$ with a proper $\delta$.
\end{remark}
\begin{remark}
Theorem \ref{def generalization} provides a potential new theoretical insight for applications incorporating compression, though some of them were not related to communication-efficient distributed training. E.g., escaping saddle point with compressed gradient \cite{avdiukhin2021escaping}, feature quantization to improve GAN training \cite{zhao2020feature}, SignSGD that empirically accelerates training \cite{bernstein2018signsgd,bernstein2018signsgd2}, etc.
\end{remark}

\begin{figure*}[t]
    \centering
    \includegraphics[width=.31\textwidth]{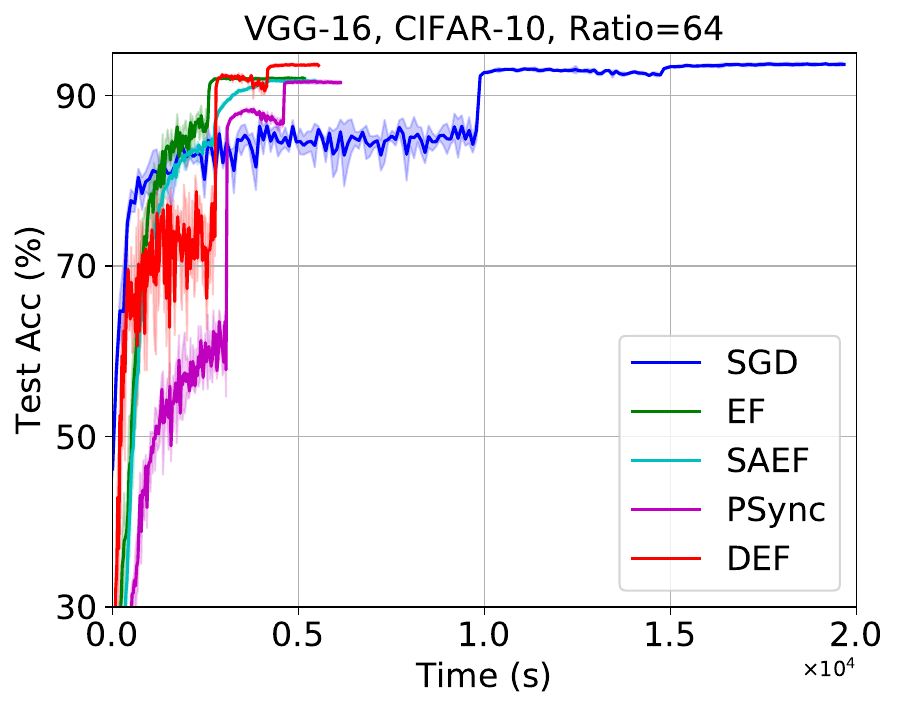}
    \includegraphics[width=.313\textwidth]{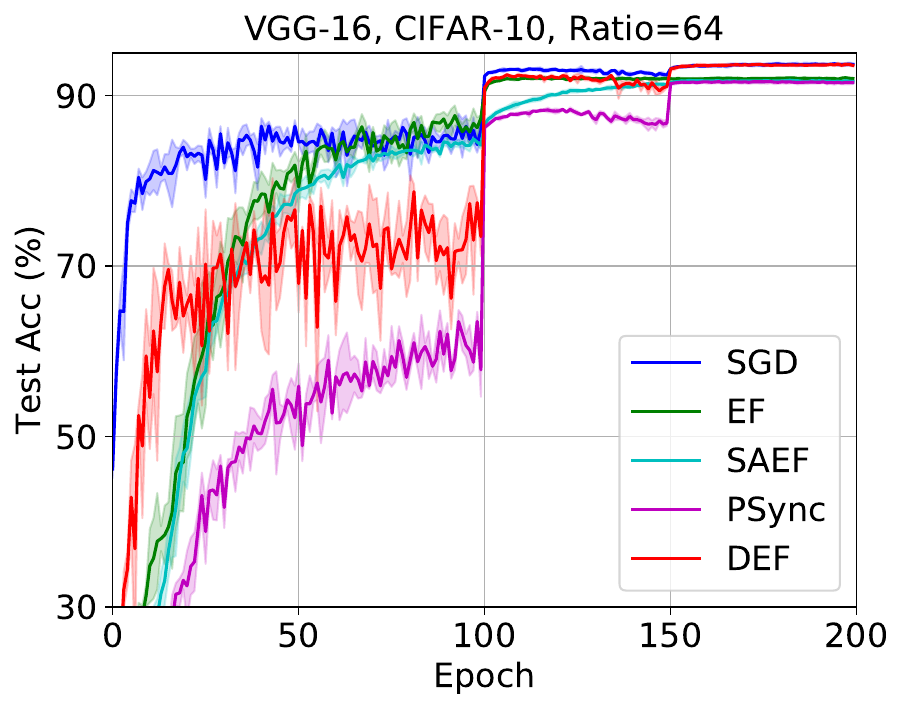}
    \includegraphics[width=.325\textwidth]{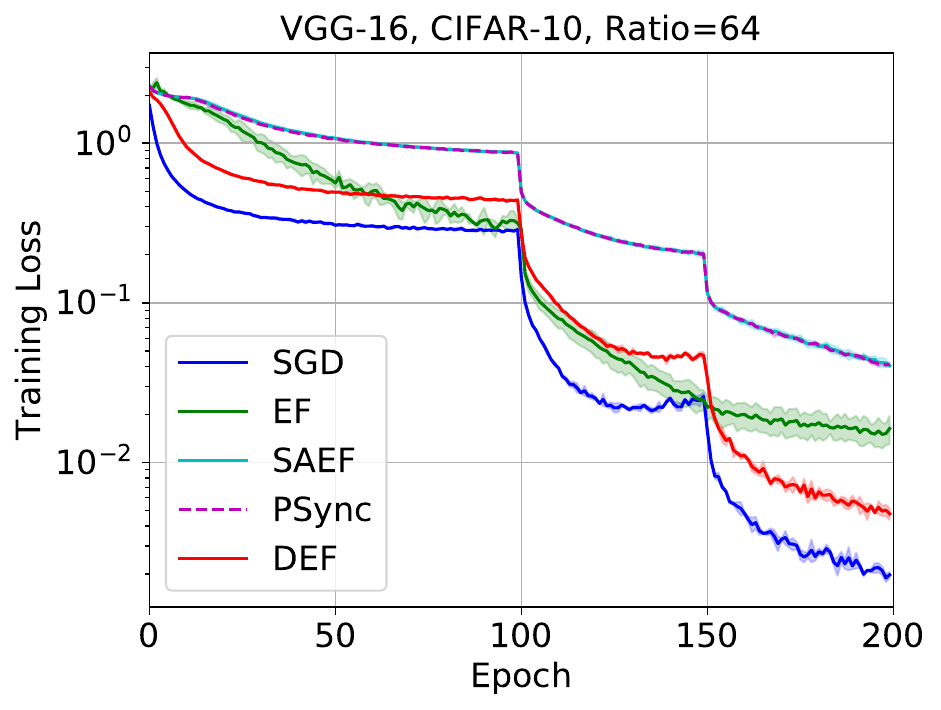}
    \includegraphics[width=.31\textwidth]{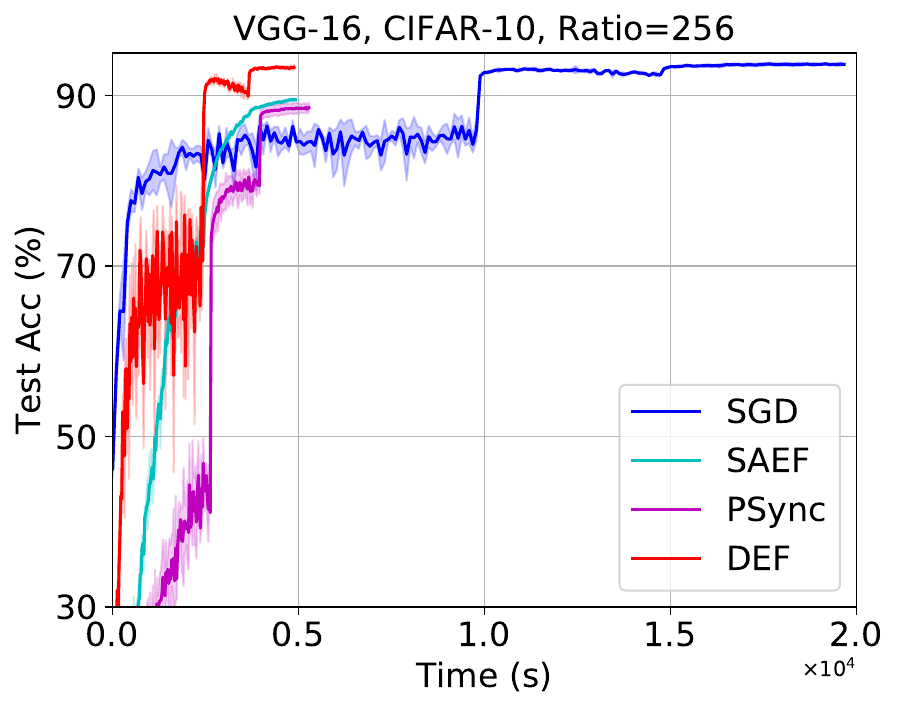}
    \includegraphics[width=.313\textwidth]{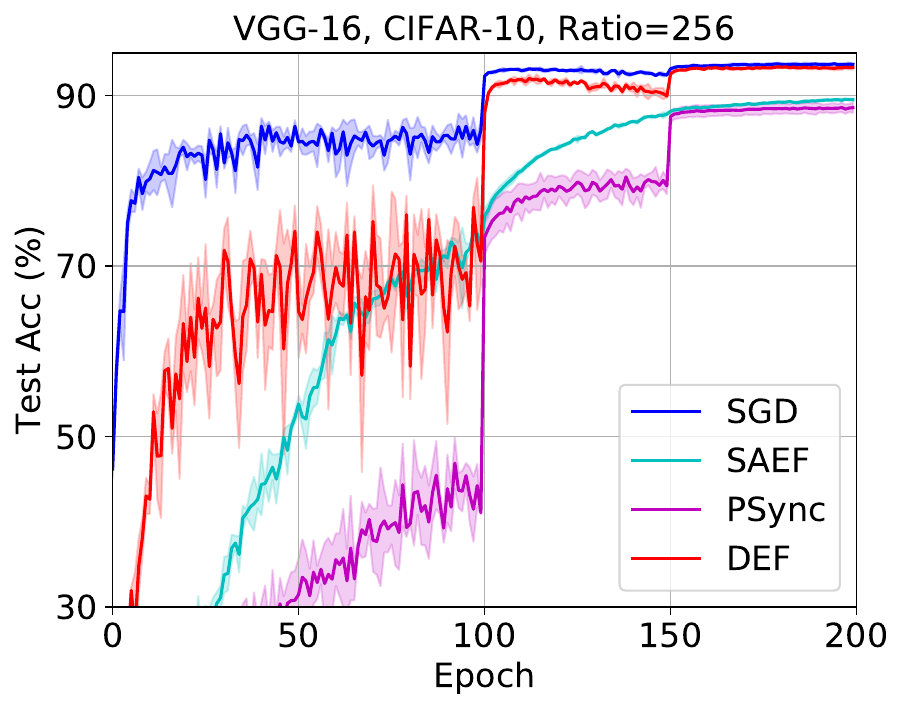}
    \includegraphics[width=.325\textwidth]{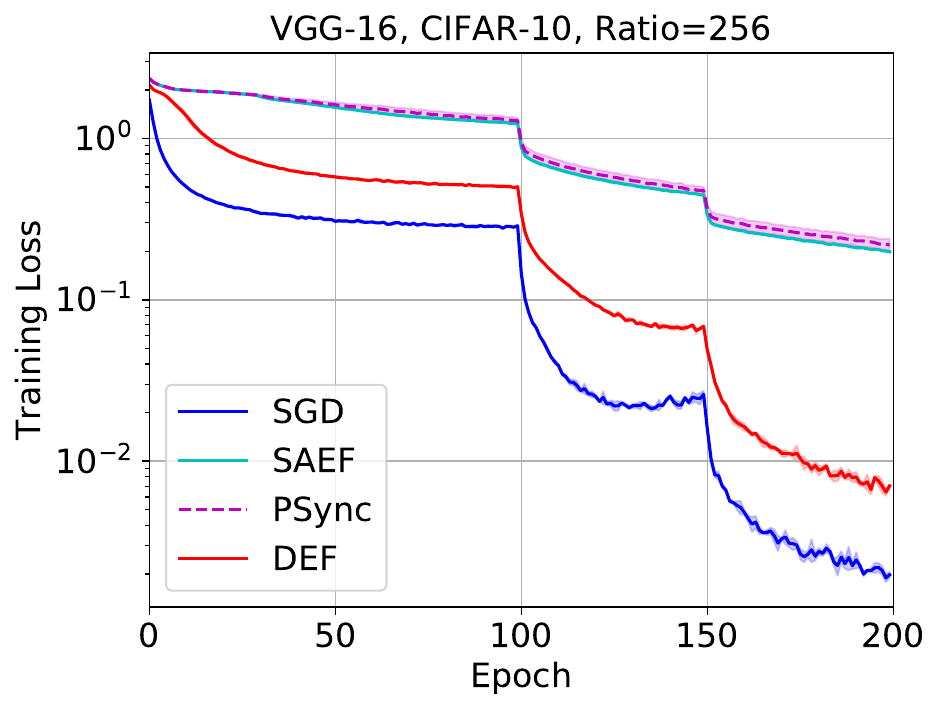}
    \caption{CIFAR-10 training curves of VGG-16. The compression ratio is 64 for the top row and 256 for the bottom row. EF is not plotted when the compression ratio is 256 due to divergence. From the left to right column, we plot the test accuracy (\%) v.s. the wall-clock time, the test accuracy (\%) v.s. training epochs, and the training loss v.s. training epochs respectively.}
    \label{fig:cifar curves}
\end{figure*}

\begin{table*}[t]
\small
    \centering
    \caption{The CIFAR-10 test accuracy (\%) comparison of under various compression ratio settings with VGG-16.}
    \label{tab:cifar acc}
    \begin{tabular}{c|c|ccc|cc}
        \toprule
        Ratio & SGD & EF & SAEF & PSync & DEF & DEF-A \\
        \midrule
        1 & 93.76 $\pm$ 0.14 & --- & --- & --- & --- & --- \\
        \midrule
        16 & --- & 93.04 $\pm$ 0.13 & 93.15 $\pm$ 0.04 & 93.31 $\pm$ 0.21 & \underline{93.61} $\pm$ 0.04 & \textbf{93.66} $\pm$ 0.10 \\
        \midrule
        64 & --- & 92.16 $\pm$ 0.06 & 91.88 $\pm$ 0.14 & 91.79 $\pm$ 0.17 & \textbf{93.75} $\pm$ 0.12 & \underline{93.61} $\pm$ 0.07 \\
        \midrule
        256 & --- & diverge & 89.59 $\pm$ 0.04 & 88.70 $\pm$ 0.61 & \textbf{93.45} $\pm$ 0.11 & \underline{93.33} $\pm$ 0.26 \\
        \midrule
        512 & --- & diverge & 87.83 $\pm$ 0.36 & 86.47 $\pm$ 0.14 & \underline{93.24} $\pm$ 0.08 & \textbf{93.25} $\pm$ 0.18 \\
        \midrule
        1024 & --- & diverge & 85.46 $\pm$ 0.80 & 84.27 $\pm$ 0.33 & \underline{93.03} $\pm$ 0.15 & \textbf{93.06} $\pm$ 0.09 \\
        \bottomrule
    \end{tabular}
\end{table*}

\begin{figure*}[t]
    \centering
    \includegraphics[width=.32\textwidth]{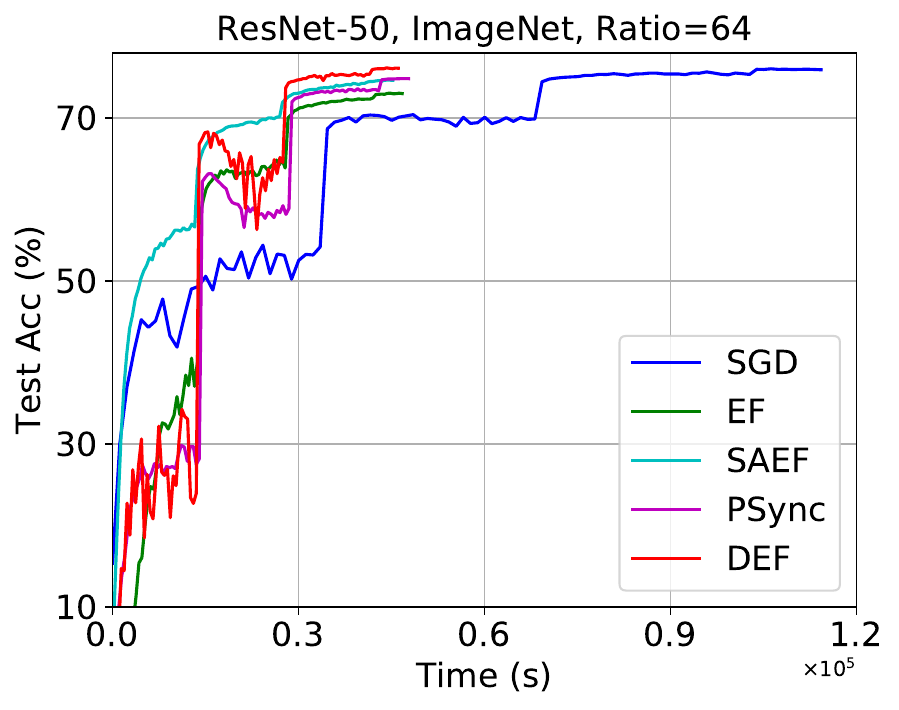}
    \includegraphics[width=.322\textwidth]{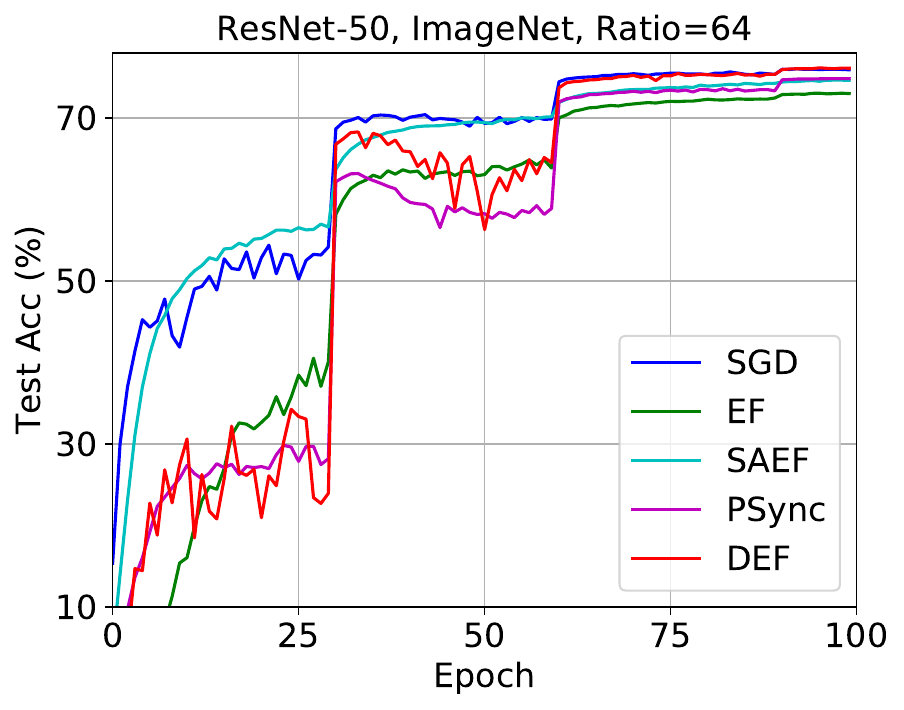}
    \includegraphics[width=.315\textwidth]{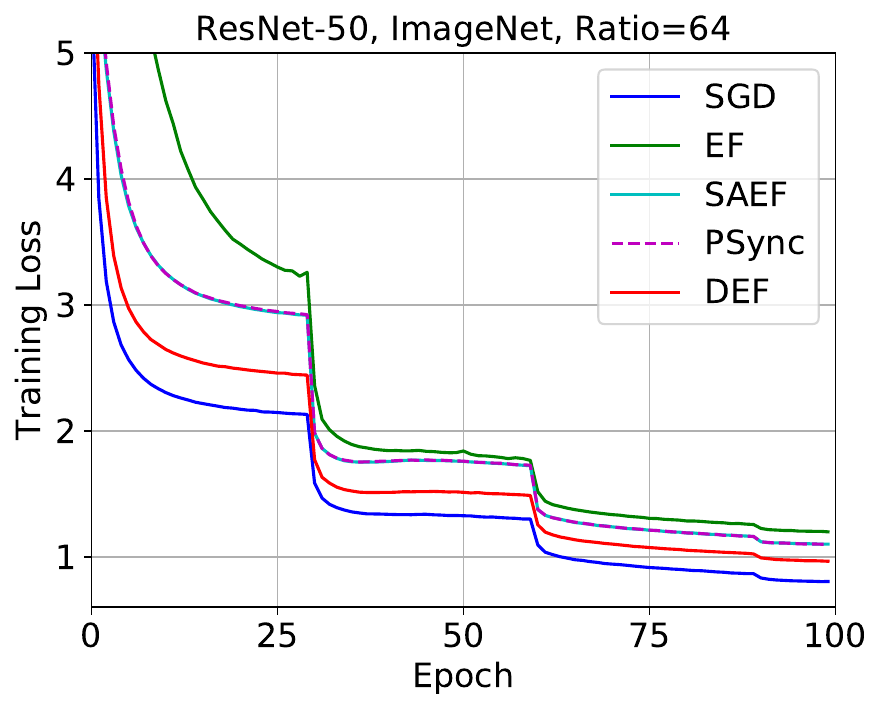}
    \includegraphics[width=.32\textwidth]{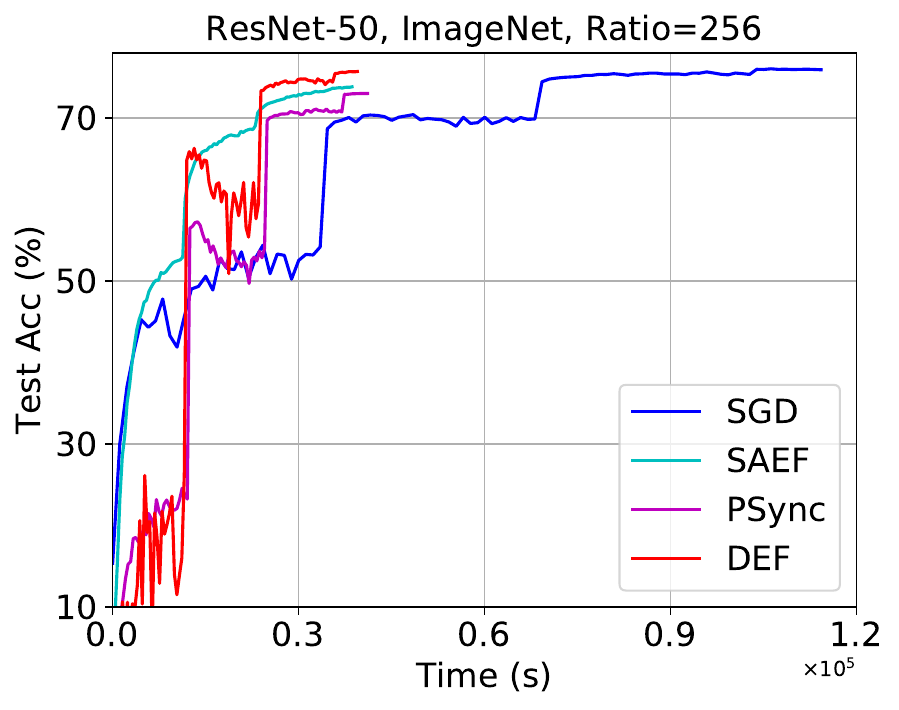}
    \includegraphics[width=.322\textwidth]{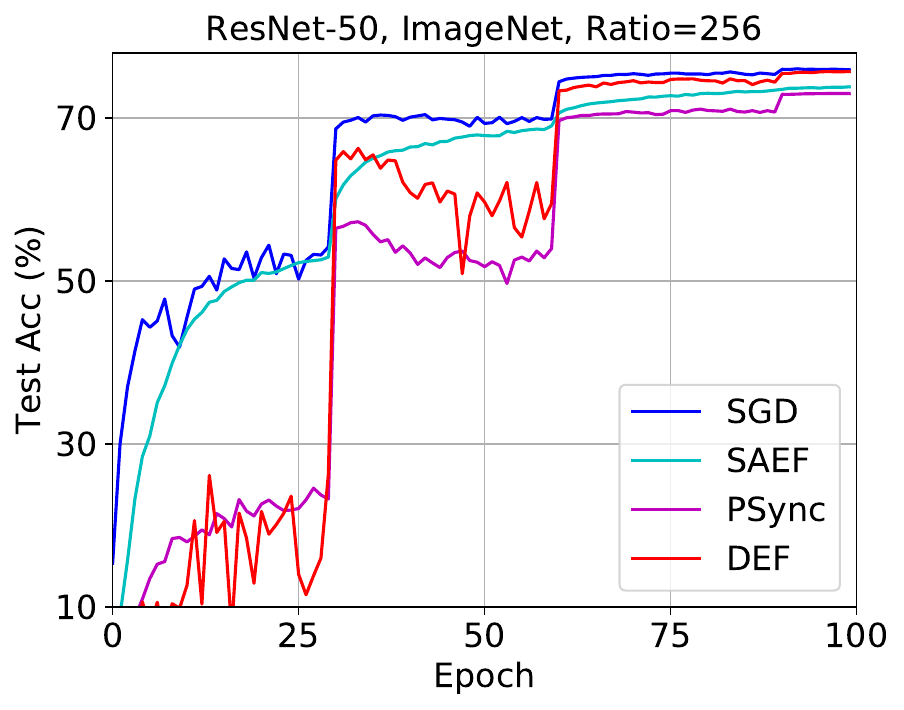}
    \includegraphics[width=.315\textwidth]{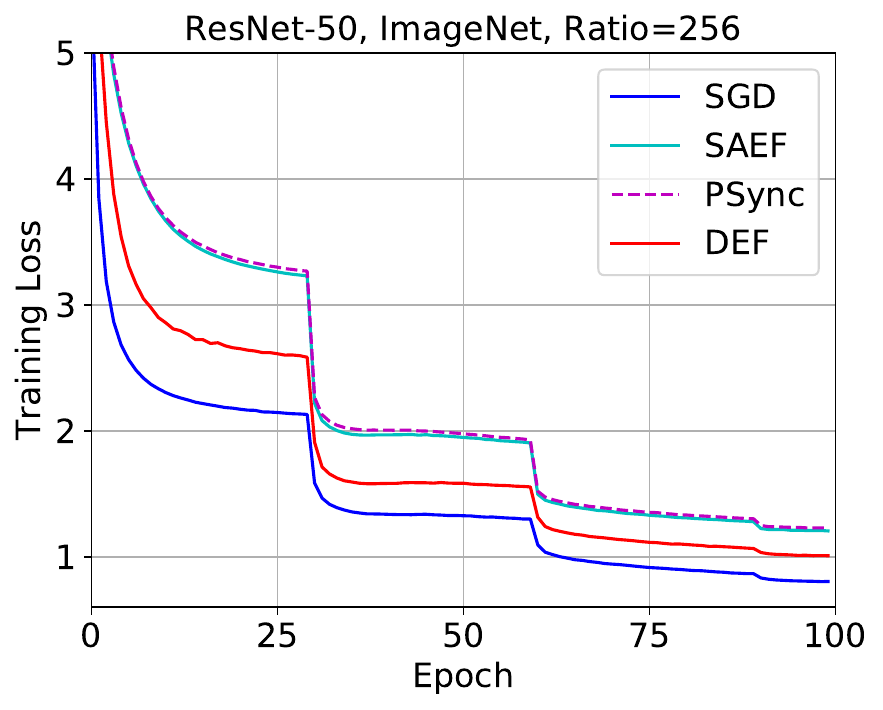}
    \caption{ImageNet training curves of ResNet-50. The compression ratio is 64 for the top row and 256 for the bottom row. From the left to right column, we plot the test accuracy (\%) v.s. the wall-clock time, the test accuracy (\%) v.s. training epochs, and the training loss v.s. training epochs respectively.}
    \label{fig:imagenet curves}
\end{figure*}

\begin{table*}[t]
\small
    \centering
    \caption{The ImageNet test accuracy (\%) comparison under various compression ratio settings with ResNet-50.}
    \label{tab:imagenet acc}
    \begin{tabular}{c|c|ccc|cc}
        \toprule
        Ratio & SGD & EF & SAEF & PSync & DEF & DEF-A \\
        \midrule
        1 & 76.04 & --- & --- & --- & --- & --- \\
        \midrule
        16 & --- & 75.29 ($\downarrow$ 0.75) & 75.83 ($\downarrow$ 0.21) & 75.63 ($\downarrow$ 0.41) & \underline{75.98} ($\downarrow$ 0.06) & \textbf{76.10} ($\uparrow$ 0.06) \\
        \midrule
        64 & --- & 73.05 ($\downarrow$ 2.99) & 74.65 ($\downarrow$ 1.39) & 74.84 ($\downarrow$ 1.20) & \underline{76.16} ($\uparrow$ 0.12) & \textbf{76.37} ($\uparrow$ 0.33) \\
        \midrule
        128 & --- & 63.80 ($\downarrow$ 12.2) & 74.26 ($\downarrow$ 1.78) & 74.12 ($\downarrow$ 1.92) & \textbf{76.17} ($\uparrow$ 0.13) & \underline{76.14}  ($\uparrow$ 0.10) \\
        \midrule
        256 & --- & diverge & 73.83 ($\downarrow$ 2.21) & 73.02 ($\downarrow$ 3.02) & \underline{75.71} ($\downarrow$ 0.33) & \textbf{76.00} ($\downarrow$ 0.04) \\
        \midrule
        512 & --- & diverge & 73.00 ($\downarrow$ 3.04) & 72.60 ($\downarrow$ 3.44) & \underline{75.52} ($\downarrow$ 0.52) & \textbf{75.77} ($\downarrow$ 0.27) \\
        \midrule
        1024 & --- & diverge & 71.89 ($\downarrow$ 4.15) & 71.82 ($\downarrow$ 4.22) & \textbf{75.64} ($\downarrow$ 0.40) & \underline{75.57} ($\downarrow$ 0.47) \\
        \bottomrule
    \end{tabular}
\end{table*}

\begin{figure}[t]
    \centering
    \includegraphics[width=.26\textwidth]{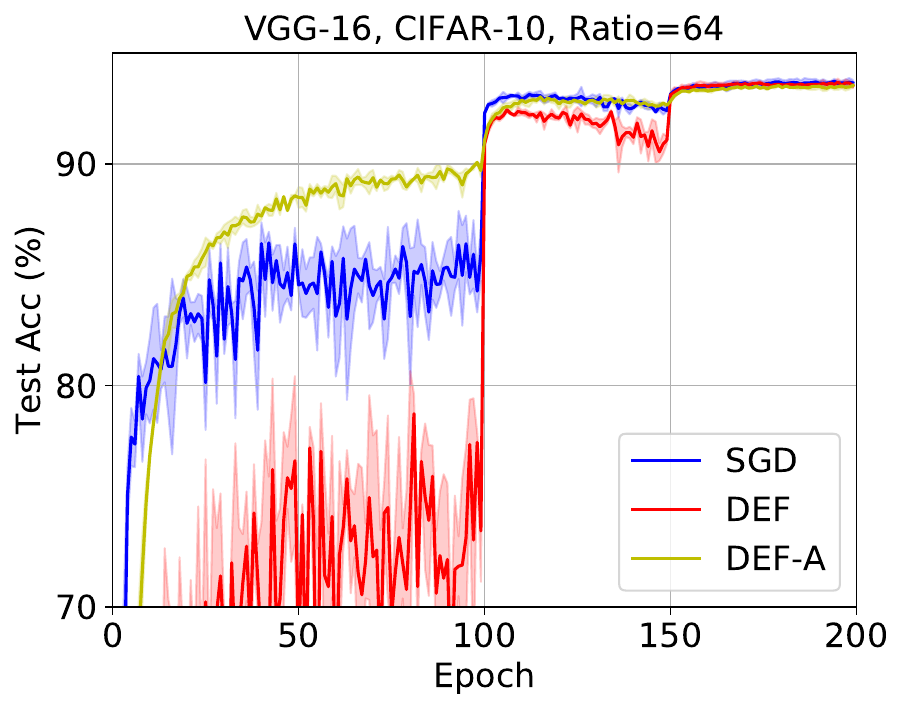}
    \includegraphics[width=.26\textwidth]{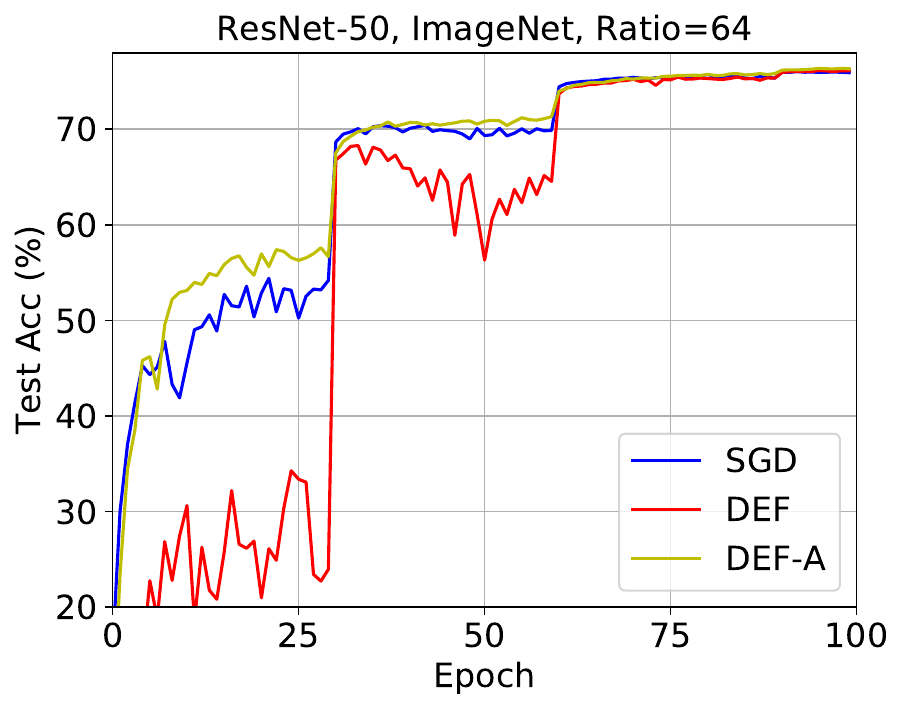}
    \caption{Accelerate the generalization with DEF-A. DEF-A significantly improves the test accuracy before the second learning rate decay compared with DEF.}
    \label{fig:error ahead}
\end{figure}

\begin{table}[t]
\small
    \centering
    \caption{The CIFAR-10/100 test accuracy (\%) comparison for various model architectures. The compression ratio is 1 for SGD and 64 for the other methods.}
    \label{tab:various model}
    \begin{tabular}{c|ccc}
        \toprule
        Method & VGG-16 & ResNet-110 & WRN-28-10 \\
        \midrule
        SGD & \makecell{93.76 $\pm$ 0.14 \\/ 72.50 $\pm$ 0.33} & \makecell{94.73 $\pm$ 0.06 \\/ 76.78 $\pm$ 0.29} & \makecell{96.21 $\pm$ 0.07 \\/ 80.81 $\pm$ 0.12} \\
        \midrule
        EF & \makecell{92.16 $\pm$ 0.06 \\/ 68.87 $\pm$ 0.21} & diverge & diverge \\
        \midrule
        SAEF & \makecell{91.88 $\pm$ 0.14 \\/ 67.00 $\pm$ 0.07} & \makecell{92.95 $\pm$ 0.17 \\/ 71.19 $\pm$ 0.31} & \makecell{95.33 $\pm$ 0.01 \\/ 79.04 $\pm$ 0.01} \\
        \midrule
        PSync & \makecell{91.79 $\pm$ 0.17 \\/ 65.68 $\pm$ 0.16} & \makecell{92.26 $\pm$ 0.04 \\/ 69.21 $\pm$ 0.04} & \makecell{95.44 $\pm$ 0.13 \\/ 79.60 $\pm$ 0.12} \\
        \midrule
        DEF & \makecell{\textbf{93.75} $\pm$ 0.12 \\/ \underline{72.02} $\pm$ 0.10} & \makecell{\underline{94.34} $\pm$ 0.06 \\/ \underline{76.43} $\pm$ 0.12} & \makecell{\textbf{96.26} $\pm$ 0.05 \\/ \underline{80.88} $\pm$ 0.16} \\
        \midrule
        DEF-A & \makecell{\underline{93.61} $\pm$ 0.07 \\/ \textbf{72.38} $\pm$ 0.07} & \makecell{\textbf{94.66} $\pm$ 0.07 \\/ \textbf{76.98} $\pm$ 0.21} & \makecell{\underline{96.24} $\pm$ 0.12 \\/ \textbf{80.95} $\pm$ 0.16} \\
        \bottomrule
    \end{tabular}
\end{table}

\begin{table}[t]
\vspace{-7pt}
\small
    \centering
    \caption{The CIFAR-10 test accuracy (\%) of DEF/DEF-A for various $\lambda$ with VGG-16. The compression ratio is 64.}
    \label{tab:various lambda}
    \begin{tabular}{c|ccc}
        \toprule
        $\lambda$ & 0.1 & 0.2 & \textbf{0.3} \\
        \midrule
        DEF & 92.83 $\pm$ 0.19 & 93.45 $\pm$ 0.06 & \textbf{93.75} $\pm$ 0.12 \\
        DEF-A & 92.78 $\pm$ 0.13 & 93.20 $\pm$ 0.14 & \underline{93.61} $\pm$ 0.07 \\
        \midrule
        \midrule
        $\lambda$ & 0.4 & 0.6 & 0.8 \\
        \midrule
        DEF & 93.41 $\pm$ 0.12 & 93.26 $\pm$ 0.10 & 92.60 $\pm$ 0.19 \\
        DEF-A & 93.41 $\pm$ 0.20 & 93.11 $\pm$ 0.20 & 92.59 $\pm$ 0.15 \\
        \bottomrule
    \end{tabular}
    \vspace{-5pt}
\end{table}

\subsection{Extension to Iterate Averaging (IA)}

As SGD and SGD-IA is a special case of DEF and DEF-A respectively when $K=1$, $\lambda=1$, and $\mathcal{C}(\Delta)=\delta\Delta$, we immediately have the following Theorem \ref{sgd-ia generalization}.
\begin{theorem}\label{sgd-ia generalization}
(Excess Risk Error of SGD(-IA), Appendix \ref{appendix:ia})  Let Assumptions \ref{bounded variance}, \ref{bounded second moment}, \ref{allreduce compressor}, \ref{lipschitz gradient} and \ref{approximate compressor} hold. Suppose $\eta=\frac{c}{t+1}$, where $c>0$ is some constant.

(1) The generalization error of SGD
\begin{equation}
    \epsilon_{\text{gen}}=\mathcal{O}(T^{(1-\frac{1}{N})Lc/((1-\frac{1}{N})Lc+1)}) \,.
\end{equation}
(2) Suppose $\eta\leq\frac{1}{4L}$. The optimization error of SGD
\begin{equation}
    \epsilon_{\text{opt}} = \widetilde{\mathcal{O}}(T^{-\frac{\mu c}{2}}+T^{-1}) \,.
\end{equation}
(3) The generalization error of SGD-IA
\begin{equation}\label{eq:sgd-ia gen}
    \epsilon_{\text{gen}}=\mathcal{O}(T^{(1-\frac{1}{N})\delta Lc/((1-\frac{1}{N})\delta Lc+1)}) \,.
\end{equation}
(4) Suppose $\eta\leq\frac{1}{8\delta L}$. The optimization error of SGD-IA
\begin{equation}
    \epsilon_{\text{opt}} = \widetilde{\mathcal{O}}(T^{-\frac{\mu\delta c}{2}} + T^{-1} + (1/\sqrt{1-\delta}-1)^{-2}) \,.
\end{equation}
\end{theorem}
\begin{remark}
We have $\delta^{\frac{1}{2}}$ in Eq.~(\ref{eq:def-a gen}) but $\delta$ in Eq.~(\ref{eq:sgd-ia gen}) because $\mathbb{E}_{\mathcal{C}}[\mathcal{C}(\Delta)]=\delta\Delta$ for RBGS but $\mathcal{C}(\Delta)=\delta\Delta$ for SGD-IA.
\end{remark}
\begin{remark}
SGD-IA can achieve better generalization rate than SGD with a proper $\delta$. \citet{neu2018iterate,wu2020obtaining} theoretically only show that SGD-IA achieves adjustable regularization for strongly-convex objective functions, while SGD-IA applications such as averaging weights \cite{izmailov2018averaging} and ensemble of models during training with cyclic learning rate \cite{huang2017snapshot} only empirically show better generalization than SGD.
\end{remark}
\begin{remark}
Compare with Theorem \ref{def generalization}, we can see that DEF-A generalizes better than SGD with a proper $\delta$.
\end{remark}
\begin{remark}
Theorem \ref{sgd-ia generalization} provides a new theoretical explanation for an important line of works in unsupervised learning - \textit{momentum contrast} \cite{he2020momentum}. In \citet{he2020momentum}, two sets of weights are maintained with a contrastive loss. One is the ``query'' $y_t$ which is updated via SGD, and the other is the ``key'' $x_t$ ($x_0=y_0$) which is updated via
\begin{equation}
    x_{t+1} = (1-\delta)x_t + \delta y_t\,.
\end{equation}
The success of momentum contrast is explained as a ``slowly progressing'' key $x_t$ \cite{he2020momentum} without theoretical guarantee. Interestingly, the above equation is identical to Eq.~(\ref{eq:special sgd-ia}), i.e. SGD-IA. Therefore, our results suggests that the slowly progressing key $x_t$ may actually have stabler and better generalization than the query $y_t$ depending on $\delta$.
\end{remark}

\section{Experiments}\label{experiment}

In this section, we conduct empirical experiments on benchmark deep learning tasks following settings in \cite{karimireddy2019error,zheng2019communication,xie2020cser} to validate the performance of the proposed detached error feedback (DEF) method. We compare the following methods with RBGS as the gradient compressor: (1) SGD, which is the upper bound without gradient compression, (2) EF \cite{karimireddy2019error,zheng2019communication}, (3) SAEF \cite{xu2021step}, (4) PSync \cite{xie2020cser}, and (5) the proposed DEF(-A), where $\lambda=0.3$ by default. We have also tested EF21 \cite{richtarik2021ef21,fatkhullin2021ef21} on our deep learning tasks with RBGS, but it does not converge.

\textbf{Settings.} All experiments are implemented using PyTorch and conducted on a cluster of machines connected by. Each machine is equipped with 4 NVIDIA P40 GPUs and there are 16 workers (GPUs) in total. We use NCCL as the backend of the PyTorch distributed package. The task-specific settings are as follows.

\textbf{CIFAR.} We train VGG-16 \cite{simonyan2014very}, ResNet-110 \cite{he2016deep} and Wide ResNet (WRN-28-10) \cite{zagoruyko2016wide} models CIFAR-10/100 \cite{krizhevsky2009learning} image classification task. We report the mean and standard deviation metrics over 3 runs. The base learning rate is tuned from $\{\cdots, 0.1, 0.05, 0.01,\cdots\}$ and the batch size is 128. The momentum constant is 0.9 and the weight decay is $5\times 10^{-4}$. The model is trained for 200 epochs with a learning rate decay of 0.1 at epoch 100 and 150. Random cropping, random flipping, and standardization are applied as data augmentation techniques.

\textbf{ImageNet.} We train the ResNet-50 model on ImageNet \cite{deng2009imagenet} image classification tasks. The model is trained for 100 epochs with a learning rate decay of 0.1 at epoch 30, 60, and 90. The base learning rate is tuned from $\{\cdots, 0.1, 0.05, 0.01,\cdots\}$ and the batch size is 256. The momentum constant is 0.9 and the weight decay is $1\times 10^{-4}$. Similar data augmentation techniques as in CIFAR experiments are applied.

\subsection{General Results}

We plot the CIFAR-10 training curves of VGG-16 in Figure \ref{fig:cifar curves} and summarize the test numbers under various compression ratio settings in Table \ref{tab:cifar acc}. From the curves, DEF achieves the best test acc and training loss among all the communication-efficient methods. Compared with SGD, DEF achieves $\times 3.6$ and $\times 4.0$ speedup when the compression ratio is 64 and 256 respectively. For the test numbers in the table, DEF and DEF-A achieve the best results, which can be comparable to SGD for compression up to 256. When the compression ratio is high, DEF(-A) can significantly improve over the best counterpart by \textbf{8\%}. Overall, DEF and DEF-A have similar final test performances. A significant improvement of the training loss over the existing EF variants can be observed, validating our lower bound in the convergence analysis of DEF.

The ImageNet training curves of ResNet-50 is shown in Figure \ref{fig:imagenet curves} and the test numbers under various compression ratio settings are summarized in Table \ref{tab:imagenet acc}. We can reach similar conclusions as in CIFAR-10 experiments. Specifically, DEF achieves $\times 2.5$ and $\times 2.9$ speedup compared with SGD when the compression ratio is 64 and 256 respectively. For the test numbers in the table, DEF and DEF-A can be comparable to SGD for the compression ratio of 1024. For some smaller compression ratios, we may even see a slight improvement over SGD. When the compression ratio is high, DEF(-A) can significantly improve the best counterpart by \textbf{4\%}.

For the concern of scalability, we also summarize the test numbers of VGG-16, ResNet-110, and WRN-28-10 on CIFAR-10/100 in Table \ref{tab:various model} with 64 as the compression ratio. DEF-A achieves lossless performance compared with SGD and largely improves all the counterparts. We find that for VGG-16 on CIFAR-100 and ResNet-110 on CIFAR-10/100, DEF-A has a noticeable improvement over DEF. In particular, we find that PSync achieves closer performance to SGD on WRN as reported in \citet{xie2020cser}, but is much worse on VGG-16 and ResNet-110. Therefore, both the superior performance and scalability of DEF(-A) are validated.

\subsection{Accelerate Generalization}

Here we empirically validate the theoretical generalization analysis that DEF-A has a better generalization rate than DEF. We plot the training curves for VGG-16 on CIFAR-10 and ResNet-50 on ImageNet with compression ratio as 64 in Figure \ref{fig:error ahead}. We can see that DEF-A does have a much faster generalization rate than DEF. Specifically, the test accuracy improvement is about \textbf{15\%} on CIFAR-10 and \textbf{25\%} on ImageNet before the first learning rate decay, which validates the theoretical benefits in our generalization analysis. DEF-A can be even faster than full-precision SGD.

A significant improvement can still be observed before the second learning rate decay, but it becomes smaller when the learning rate is smaller. This matches our generalization analysis well. Let $c$ be smaller such that the learning rate $\eta=\frac{c}{t+1}$ is smaller, then Eq.~(\ref{eq:def gen}) is closer to Eq.~(\ref{eq:def-a gen}), that is, the DEF-A's generalization error improvement over DEF becomes smaller. Then it is obvious that the excess risk error improvement will also become smaller.

\subsection{Hyper-parameter $\lambda$}
Here we explore DEF(-A) with various choices of the hyper-parameter $\lambda$ with results summarized in Table \ref{tab:various lambda}. We can just set $\lambda=0.3$ by default for the best performance. In comparison, an inappropriate choice of $\lambda$ (e.g., 0.1 and 0.8) can lead to the performance degradation of about 1\%. We also observe that a wide range of $\lambda$ such as 0.2 $\sim$ 0.6 can result in fairly good performance compared with $\lambda=0.3$, which means that the proposed DEF(-A) is not too sensitive to the hyper-parameter $\lambda$.



\section{Conclusion}

In this work, to address the performance loss issue for communication-efficient distributed SGD with the gradient sparsifier RBGS, we proposed a new DEF(-A) algorithm motivated by the trade-off between gradient variance and second moment. Our convergence analysis shows better bounds without relying on the bound of gradient second moment. We conduct the first generalization analysis for communication-efficient distributed training to show that DEF-A can generalize faster than DEF and SGD, which sheds light on other applications incorporating compression such as escaping saddle point, GAN training, and SignSGD training. We establish the connection to SGD-IA for the first time, thus our analysis provides potential theoretical explanations for SGD-IA applications such as averaging weights, ensemble, and momentum contrast in unsupervised learning. Last but not least, deep learning experiments validate the significant improvement of DEF(-A) over existing EF variants.

\bibliography{example_paper}
\bibliographystyle{icml2022}

\newpage
\input{appendix_v2}


\end{document}

%% file: appendix_v2.tex
\appendix
\onecolumn

\section{Proof of Convergence of DEF (Theorem \ref{def convergence})}\label{appendix:convergence}
In this section, we consider general non-convex objective functions with $\delta$-approximate and ring-allreduce compatible compressor. We want the following term to be as close to zero as possible.
\begin{equation}
\begin{split}
    \frac{1}{K}\sum^{K}_{k=1}\|\frac{1}{K}\sum^{K}_{k^\prime=1}e_{k^\prime,t} - \lambda_t e_{k,t}\|^2_2 \,.
\end{split}
\end{equation}
For ease of notation, let $e_t\coloneqq \frac{1}{K}\sum^{K}_{k=1}e_{k,t}$, $\text{Var}(e_t)\coloneqq \frac{1}{K}\sum^{K}_{k=1}\|e_{k,t}-e_t\|^2_2$. We have $\frac{1}{K}\sum^{K}_{k=1}\|e_{k,t}\|^2_2=\|e_t\|^2_2 + \text{Var}(e_t)$. Then
\begin{equation}
\begin{split}
    \frac{1}{K}\sum^{K}_{k=1}\|\frac{1}{K}\sum^{K}_{k^\prime=1}e_{k,t} - \lambda_t e_{k,t}\|^2_2 = \frac{1}{K}\sum^{K}_{k=1}\|e_t - \lambda_t e_{k,t}\|^2 &= \frac{1}{K}\sum^{K}_{k=1}(\|e_t\|^2_2 - 2\lambda_t\langle e_t, e_{k,t}\rangle + \lambda_t^2\|e_{k,t}\|^2_2) \\
    &= \frac{1}{K}\sum^{K}_{k=1}\|e_{k,t}\|^2_2 \lambda_t^2 - 2\|e_t\|^2_2 \lambda_t + \|e_t\|^2_2\\
    &= (\|e_t\|^2_2 + \text{Var}(e_t))\lambda_t^2 - 2\|e_t\|^2_2 \lambda_t + \|e_t\|^2_2 \,.\\
\end{split}
\end{equation}
When $\lambda_t^*=\frac{\|e_t\|^2_2}{\|e_t\|^2_2 + \text{Var}(e_t)}$, it has the minimum $\frac{\|e_t\|^2_2 \cdot \text{Var}(e_t)}{\|e_t\|^2_2 + \text{Var}(e_t)}$.

If we allow individual coefficient $\lambda_{k,t}$ for each worker $k\in[K]$, then $\lambda_{k,t}^*=\frac{\langle e_t,e_{k,t}\rangle}{\|e_{k,t}\|^2_2}$ by minimizing $\|e_t-\lambda_{k,t}e_{k,t}\|^2_2$. However, it is impractical due to the additional $K$ hyper-parameters to tune when $K$ is large.

For simplicity, let $\lambda_t\rightarrow \lambda$ because it is hard to manually tune $\lambda_t$ during the training.

\subsection{Lemmas}
For ease of notation, let $g_{k,t}$ denotes the stochastic gradient computed at iteration $t$ for worker $k$ and $g_t\coloneqq \frac{1}{K}\sum^{K}_{k=1}g_{k,t}$.

\begin{lemma}\label{lemma:grad diff} Let Assumptions \ref{bounded variance}, \ref{bounded second moment}, and \ref{lipschitz gradient} hold. Let $B_1=(\frac{1}{K}-\lambda)^2+\frac{K-1}{K^2}$ and $B_2=|\frac{1}{K}-\lambda|+\frac{K-1}{K}$. We have
\begin{equation}
\begin{split}
    \frac{1}{K}\sum^{K}_{k=1}\mathbb{E}\|g_t - \lambda g_{k,t}\|^2_2 \leq B_1\sigma^2 + 2(1-\lambda)^2M^2 + 2B_2^2L^2\cdot\frac{1}{K}\sum^{K}_{k=1}\mathbb{E}\|e_t-\lambda e_{k,t}\|^2_2 \,,
\end{split}
\end{equation}
\end{lemma}
\begin{proof}
We know that $g_{k,t}=\nabla f(x_t-\lambda e_{k,t};\xi_{k,t})$. Then
\begin{equation}
\begin{split}
    &\frac{1}{K}\sum^{K}_{k=1}\mathbb{E}\|g_t - \lambda g_{k,t}\|^2_2 \\
    &= \frac{1}{K}\sum^{K}_{k=1}\mathbb{E}\|\frac{1}{K}\sum^{K}_{k^\prime=1}[\nabla f(x_t-\lambda e_{k^\prime,t};\xi_{k^\prime,t}) - \nabla F_{\mathcal{S}}(x_t-\lambda e_{k^\prime,t})] - \lambda [\nabla f(x_t-\lambda e_{k,t};\xi_{k,t}) - \nabla F_{\mathcal{S}}(x_t-\lambda e_{k,t})] \\
    &\quad\quad\quad\quad\quad + \frac{1}{K}\sum^{K}_{k^\prime=1}[\nabla F_{\mathcal{S}}(x_t-\lambda e_{k^\prime,t}) - \nabla F_{\mathcal{S}}(y_t)] - \lambda [\nabla F_{\mathcal{S}}(x_t-\lambda e_{k,t}) - \nabla F_{\mathcal{S}}(y_t)] + (1-\lambda)\nabla F_{\mathcal{S}}(y_t)\|^2_2 \\
    &\overset{(a)}{=} \underbrace{\frac{1}{K}\sum^{K}_{k=1}\mathbb{E}\|\frac{1}{K}\sum^{K}_{k^\prime=1}[\nabla f(x_t-\lambda e_{k^\prime,t};\xi_{k^\prime,t}) - \nabla F_{\mathcal{S}}(x_t-\lambda e_{k^\prime,t})] - \lambda [\nabla f(x_t-\lambda e_{k,t};\xi_{k,t}) - \nabla F_{\mathcal{S}}(x_t-\lambda e_{k,t})] \|^2_2}_{\tcircle{1}} \\
    &\quad + \frac{1}{K}\sum^{K}_{k=1}\mathbb{E}\|\frac{1}{K}\sum^{K}_{k^\prime=1}[\nabla F_{\mathcal{S}}(x_t-\lambda e_{k^\prime,t}) - \nabla F_{\mathcal{S}}(y_t)] - \lambda [\nabla F_{\mathcal{S}}(x_t-\lambda e_{k,t}) - \nabla F_{\mathcal{S}}(y_t)] + (1-\lambda)\nabla F_{\mathcal{S}}(y_t)\|^2_2 \\
    &\overset{(b)}{\leq} \tcircle{1} + \underbrace{\frac{2}{K}\sum^{K}_{k=1}\mathbb{E}\|\frac{1}{K}\sum^{K}_{k^\prime=1}[\nabla F_{\mathcal{S}}(x_t-\lambda e_{k^\prime,t}) - \nabla F_{\mathcal{S}}(y_t)] - \lambda [\nabla F_{\mathcal{S}}(x_t-\lambda e_{k,t}) - \nabla F_{\mathcal{S}}(y_t)]\|^2_2}_{\tcircle{2}} + 2(1-\lambda)^2M^2 \,,
\end{split}
\end{equation}
where (a) is due to $\nabla F_{\mathcal{S}}(x_t-\lambda e_{k,t})=\mathbb{E}g_{k,t}=\mathbb{E}\nabla f(x_t-\lambda e_{k,t};\xi_{k,t})$, (b) follows Assumption \ref{bounded second moment}. Now we consider term \tcircle{1}. For simplicity, let $a_k=\nabla f(x_t-\lambda e_{k,t};\xi_{k,t}) - \nabla F_{\mathcal{S}}(x_t-\lambda e_{k,t})$ and we will have $\mathbb{E}\langle a_k,a_{k^\prime}\rangle=0$ when $k\neq k^\prime$. Then
\begin{equation}
\begin{split}
    &\tcircle{1} = \frac{1}{K}\sum^{K}_{k=1}\mathbb{E}\|(\frac{1}{K}-\lambda)a_k + \frac{1}{K}\sum^{K}_{k^\prime=1,k^\prime\neq k}a_{k^\prime}\|^2_2 = \frac{1}{K}\sum^{K}_{k=1}\underbrace{[(\frac{1}{K}-\lambda)^2+\frac{K-1}{K^2}]}_{B_1}\|a_k\|^2_2 \overset{(a)}{\leq} B_1\sigma^2 \,,
\end{split}
\end{equation}
where (a) follows Assumption \ref{bounded variance}. $B_1=0$ if and only if $K=1$ and $\lambda=1$. Now we consider term \tcircle{2}. For simplicity, let $b_k=\nabla F_{\mathcal{S}}(x_t-\lambda e_{k,t}) - \nabla F_{\mathcal{S}}(y_t)$ and $B_2=|\frac{1}{K}-\lambda|+\frac{K-1}{K}$. $B_2=0$ if and only if $K=1$ and $\lambda=1$. Then
\begin{equation}
\begin{split}
    &\tcircle{2} = \frac{2}{K}\sum^{K}_{k=1}\mathbb{E}\|\frac{1}{K}\sum^{K}_{k^\prime=1}b_{k^\prime}-\lambda b_k\|^2_2 = \frac{2}{K}\sum^{K}_{k=1}\mathbb{E}\|(\frac{1}{K}-\lambda)b_k+\frac{1}{K}\sum^{K}_{k^\prime=1,k^\prime\neq k}b_{k^\prime}\|^2_2 \\
    &= \frac{2}{K}\sum^{K}_{k=1}B_2^2\mathbb{E}\|\frac{\frac{1}{K}-\lambda}{B_2}b_k + \frac{1}{KB_2}\sum^{K}_{k^\prime=1,k^\prime\neq k}b_{k^\prime}\|^2_2 \\
    &\leq \frac{2}{K}\sum^{K}_{k=1}B_2^2\mathbb{E}(\frac{|\frac{1}{K}-\lambda|}{B_2}\|b_k\|^2_2 + \frac{1}{KB_2}\sum^{K}_{k^\prime=1,k^\prime\neq k}\|b_{k^\prime}\|^2_2) \\
    &=\frac{2}{K}\sum^{K}_{k=1}B_2^2(\frac{|\frac{1}{K}-\lambda|}{B_2}+\frac{K-1}{KB_2})\mathbb{E}\|b_k\|^2_2 = \frac{2B_2^2}{K}\sum^{K}_{k=1}\mathbb{E}\|b_k\|^2_2 \\
    &\overset{(a)}{\leq} 2B_2^2L^2\cdot\frac{1}{K}\sum^{K}_{k=1}\mathbb{E}\|e_t-\lambda e_{k,t}\|^2_2 \,,
\end{split}
\end{equation}
where (a) follows Assumption \ref{lipschitz gradient}. Substitute terms \tcircle{1} and \tcircle{2} with their bounds and we can complete the proof.

\end{proof}

\begin{lemma}\label{lemma:error diff}
Let Assumptions \ref{bounded variance}, \ref{bounded second moment}, \ref{lipschitz gradient}, \ref{approximate compressor} and \ref{allreduce compressor} hold. Let $B_1=(\frac{1}{K}-\lambda)^2+\frac{K-1}{K^2}$ and $\eta<\frac{1}{2L}$, we have
\begin{equation}
    \frac{1}{K}\sum^{K}_{k=1}\mathbb{E}\|\frac{1}{K}\sum^{K}_{k^\prime=1}e_{k^\prime,t} - \lambda e_{k,t}\|^2_2 \leq \frac{1}{(\sqrt{\frac{1-\delta/2}{1-\delta}}-1)^2}\eta^2[B_1\sigma^2 + 2(1-\lambda)^2M^2]\,.
\end{equation}
\end{lemma}
\begin{proof}
We have
\begin{equation}
\begin{split}
    &\frac{1}{K}\sum^{K}_{k=1}\mathbb{E}\|\frac{1}{K}\sum^{K}_{k^\prime=1}e_{k^\prime,t} - \lambda e_{k,t}\|^2_2 = \frac{1}{K}\sum^{K}_{k=1}\mathbb{E}\|e_t - \lambda e_{k,t}\|^2_2 \\
    &\overset{(a)}{=} \frac{1}{K}\sum^{K}_{k=1}\mathbb{E}\|(\eta g_{t-1}+e_{t-1})-\mathcal{C}(\eta g_{t-1}+e_{t-1}) - \lambda (\eta g_{k,t-1}+e_{k,t-1}) + \lambda \mathcal{C}(\eta g_{k,t-1}+e_{k,t-1})\|^2_2 \\
    &\overset{(a)}{=} \frac{1}{K}\sum^{K}_{k=1}\mathbb{E}\|(\eta g_{t-1}-\lambda\eta g_{k,t-1}+e_{t-1}-\lambda e_{k,t-1}) - \mathcal{C}(\eta g_{t-1}-\lambda\eta g_{k,t-1}+e_{t-1}-\lambda e_{k,t-1})\|^2_2 \\
    &\overset{(b)}{=} (1-\delta)\cdot\frac{1}{K}\sum^{K}_{k=1}\mathbb{E}\|\eta g_{t-1} - \lambda \eta g_{k,t-1} + e_{t-1} - \lambda e_{k,t-1}\|^2_2 \\
    &\leq (1-\delta)(1+\beta)\cdot\frac{1}{K}\sum^{K}_{k=1}\mathbb{E}\|e_{t-1}-\lambda e_{k,t-1}\|^2_2 + (1-\delta)(1+\frac{1}{\beta})\eta^2\cdot\frac{1}{K}\sum^{K}_{k=1}\mathbb{E}\|g_{t-1}-\lambda g_{k,t-1}\|^2_2 \\
    &\overset{(c)}{=} (1-\delta)(1+\beta)\cdot\frac{1}{K}\sum^{K}_{k=1}\mathbb{E}\|e_{t-1}-\lambda e_{k,t-1}\|^2_2 + (1-\delta)(1+\frac{1}{\beta})\eta^2[B_1\sigma^2+2(1-\lambda)^2M^2] \\
    &\quad + (1-\delta)(1+\frac{1}{\beta})2B_2^2\eta^2L^2 \cdot \frac{1}{K}\sum^{K}_{k=1}\mathbb{E}\|e_{t-1}-\lambda e_{k,t-1}\|^2_2 \,,
\end{split}
\end{equation}
where (a) follows Assumption \ref{allreduce compressor}, (b) follows Assumption \ref{approximate compressor}, and (c) follows Lemma \ref{lemma:grad diff}. $\beta$ is a constant such that $0<\beta<\frac{\delta}{1-\delta}$, i.e., $(1-\delta)(1+\beta)<1$. Let $B_3=(1-\delta)(1+\frac{1}{\beta})2B_2^2\eta^2L^2 < 1 - (1-\delta)(1+\beta)$, i.e., $B_3+(1-\delta)(1+\beta)<1$, then
\begin{equation}
\begin{split}
    &\frac{1}{K}\sum^{K}_{k=1}\mathbb{E}\|e_t-\lambda e_{k,t}\|^2_2\\
    &\leq [B_3+(1-\delta)(1+\beta)]\frac{1}{K}\sum^{K}_{k=1}\mathbb{E}\|e_{t-1}-\lambda e_{k,t-1}\|^2_2 + (1-\delta)(1+\frac{1}{\beta})\eta^2[B_1\sigma^2 + 2(1-\lambda)^2M^2]\\
    &= (1-\delta)(1+\frac{1}{\beta})[B_1\sigma^2 + 2(1-\lambda)^2M^2]\sum^{t-1}_{t^\prime=0}[B_3+(1-\delta)(1+\beta)]^{t-1-t^\prime}\eta^2 \\
    &< \underbrace{\frac{(1-\delta)(1+\frac{1}{\beta})}{1 - B_3 - (1-\delta)(1+\beta)}}_{h(\beta)} \eta^2[B_1\sigma^2 + 2(1-\lambda)^2M^2] \,.\\
\end{split}
\end{equation}
Now we consider the minimum value of $h(\beta)$. Its gradient regarding $\beta$ is
\begin{equation}
\begin{split}
    \frac{\partial h(\beta)}{\partial \beta} = \frac{1-\delta}{\beta^2[1-B_3-(1-\delta)(1+\beta)]^2}[(1-\delta)\beta^2+2(1-\delta)\beta + B_3-\delta]\,.
\end{split}
\end{equation}
Therefore,
\begin{equation}
\begin{split}
    \beta^* &= -1 + \sqrt{\frac{1-B_3}{1-\delta}} \rightarrow B_3=1 - (1-\delta)(1+\beta^*)^2 < 1-(1-\delta)(1+\beta^*) \text{, valid,} \\
    h(\beta^*) &= \frac{1-\delta}{(\sqrt{1-B_3}-\sqrt{1-\delta})^2} = \frac{1}{(\sqrt{\frac{1-B_3}{1-\delta}}-1)^2} \,,\\
    \frac{1}{K}\sum^{K}_{k=1}\mathbb{E}\|e_t-\lambda e_{k,t}\|^2_2 &\leq \frac{1}{(\sqrt{\frac{1-B_3}{1-\delta}}-1)^2}\eta^2[B_1\sigma^2 + 2(1-\lambda)^2M^2] \,,\\
\end{split}
\end{equation}
which completes the proof. For simplicity we can just set $B_3\leq\delta/2$ (we can choose a constant $>1$ other than 2), which is valid as it leads to $\beta^*\leq -1+\sqrt{\frac{1-\delta/2}{1-\delta}}$ and $-1+\sqrt{\frac{1-\delta/2}{1-\delta}}<\frac{\delta}{1-\delta}$ holds. Based on the definition of $B_3$, it also requires that
\begin{equation}
    2B_2^2\eta^2L^2 < \frac{\delta/2}{(1-\delta)(-1+\sqrt{\frac{1-\delta/2}{1-\delta}})} = \frac{\delta/2}{-(1-\delta)+\sqrt{(1-\delta)(1-\delta/2)}}\,,
\end{equation}
where the R.H.S. is monotonically increasing for $0<\delta<1$. Therefore, for all conditions above to hold, we only need to assume that
\begin{equation}
    2B_2^2\eta^2L^2 \leq 4(1+(1-\lambda)^2)\eta^2L^2<\lim_{\delta\rightarrow 0}\frac{\delta/2}{-(1-\delta)+\sqrt{(1-\delta)(1-\delta/2)}}=2\,.
\end{equation}
As $0<\lambda<1$, we can simply assume $\eta<\frac{1}{2L}$.
\end{proof}

\begin{lemma}\label{lemma:error bound}
Let Assumptions \ref{bounded variance}, \ref{bounded second moment}, \ref{approximate compressor}, and \ref{allreduce compressor} hold. We have $\frac{1}{K}\sum^{K}_{k=1}\mathbb{E}\|e_{k,t}\|^2_2 \leq \frac{\eta^2(\sigma^2+M^2)}{(\sqrt{1/(1-\delta)}-1)^2}$.
\end{lemma}
\begin{proof}
We have
\begin{equation}
\begin{split}
    &\frac{1}{K}\sum^{K}_{k=1}\mathbb{E}\|e_{k,t}\|^2_2\overset{(a)}{=} \frac{1}{K}\sum^{K}_{k=1}\mathbb{E}\|\eta g_{k,t-1}+e_{k,t-1} - \mathcal{C}(\eta g_{k,t-1}+e_{k,t-1})\|^2_2 \\
    &\overset{(b)}{\leq} \frac{1-\delta}{K}\sum^{K}_{k=1}\mathbb{E}\|\eta g_{k,t-1}+e_{k,t-1}\|^2_2 \\
    &\leq (1-\delta)(1+\beta)\cdot\frac{1}{K}\sum^{K}_{k=1}\mathbb{E}\|e_{k,t-1}\|^2_2 + (1-\delta)(1+\frac{1}{\beta})\eta^2(\sigma^2+M^2) \\
    &\leq (1-\delta)(1+\frac{1}{\beta})(\sigma^2+M^2)\sum^{t-1}_{t^\prime=0}[(1-\delta)(1+\beta)]^{t-1-t^\prime}\eta^2 \\
    &\leq \frac{(1-\delta)(1+\frac{1}{\beta})}{1-(1-\delta)(1+\beta)}\eta^2(\sigma^2+M^2) \\
    &= \frac{1}{(\sqrt{1/(1-\delta)}-1)^2}\eta^2(\sigma^2+M^2) \,, \\
\end{split}
\end{equation}
where $\beta=-1+\frac{1}{\sqrt{1-\delta}}$, (a) follows Assumption \ref{allreduce compressor}, and (b) follows Assumption \ref{approximate compressor}.
\end{proof}

\subsection{Main Proof}
In this section, we need Assumptions \ref{bounded variance}, \ref{bounded second moment}, \ref{lipschitz gradient}, \ref{approximate compressor}, \ref{allreduce compressor}, and $\eta\leq \frac{1}{4L}$.

Firstly, we have the update rule of $y_t$
\begin{equation}
\begin{split}
    &y_{t+1} = x_{t+1} - e_{t+1} = x_{t+1} - \frac{1}{K}\sum^{K}_{k=1}e_{k,t+1} \\
    &= x_{t} - \frac{1}{K}\sum^{K}_{k=1}\mathcal{C}(\eta g_{k,t} + e_{k,t}) - \frac{1}{K}\sum^{K}_{k=1}(\eta g_{k,t} + e_{k,t} - \mathcal{C}(\eta g_{k,t} + e_{k,t})) \\
    &=x_t - \frac{1}{K}\sum^{K}_{k=1}(\eta g_{k,t}+e_{k,t}) = y_t - \frac{1}{K}\sum^{K}_{k=1}\eta g_{k,t} = y_t - \eta g_t \,.
\end{split}
\end{equation}
According to the Liptschitz gradient assumption,
\begin{equation}
\begin{split}
    &\mathbb{E}[F_{\mathcal{S}}(y_{t+1}) - F_{\mathcal{S}}(y_t)] \leq \mathbb{E}\langle\nabla F_{\mathcal{S}}(y_t), y_{t+1} - y_t\rangle + \frac{L}{2}\mathbb{E}\|y_{t+1} - y_t\|^2_2 \\
    &= \mathbb{E}\langle\nabla F_{\mathcal{S}}(y_t),-\frac{\eta}{K}\sum^{K}_{k=1}g_{k,t})\rangle + \frac{\eta^2L}{2}\mathbb{E}\|\frac{1}{K}\sum^{K}_{k=1}g_{k,t}\|^2_2 \\
    &=\underbrace{-\frac{\eta}{K}\sum^{K}_{k=1}\mathbb{E}\langle\nabla F_{\mathcal{S}}(y_t), \nabla F_{\mathcal{S}}(x_t-\lambda e_{k,t})\rangle}_{\tcircle{1}} + \frac{\eta^2L}{2}\underbrace{\frac{1}{K}\sum^{K}_{k=1}\mathbb{E}\|\nabla F_{\mathcal{S}}(x_t-\lambda e_{k,t})\|^2_2}_{\tcircle{2}} + \frac{\eta^2L\sigma^2}{2K} \,.
\end{split}
\end{equation}
For term \tcircle{1}, we have
\begin{equation}
\begin{split}
    &\tcircle{1} = -\eta\mathbb{E}\|\nabla F_{\mathcal{S}}(y_t)\|^2_2 - \frac{\eta}{K}\sum^{K}_{k=1}\mathbb{E}\langle\nabla F_{\mathcal{S}}(y_t),\nabla F_{\mathcal{S}}(x_t-\lambda e_{k,t})-\nabla F_{\mathcal{S}}(y_t)\rangle \\
    &\leq -\frac{\eta}{2}\mathbb{E}\|\nabla F_{\mathcal{S}}(y_t)\|^2_2 + \frac{\eta}{2K}\sum^{K}_{k=1}\mathbb{E}\|\nabla F_{\mathcal{S}}(x_t-\lambda e_{k,t})-\nabla F_{\mathcal{S}}(y_t)\|^2_2 \\
    &\leq -\frac{\eta}{2}\mathbb{E}\|\nabla F_{\mathcal{S}}(y_t)\|^2_2 + \frac{\eta L^2}{2K}\sum^{K}_{k=1}\mathbb{E}\|e_t-\lambda e_{k,t}\|^2_2 \,.
\end{split}
\end{equation}
For term \tcircle{2}, we have
\begin{equation}
\begin{split}
    &\tcircle{2} \leq \frac{1}{K}\sum^{K}_{k=1}\mathbb{E}[2\|\nabla F_{\mathcal{S}}(x_t-\lambda e_{k,t}) - \nabla F_{\mathcal{S}}(y_t)\|^2_2 + 2\|\nabla F_{\mathcal{S}}(y_t)\|^2_2] \\
    &\leq 2\mathbb{E}\|\nabla F_{\mathcal{S}}(y_t)\|^2_2 + \frac{2L^2}{K}\sum^{K}_{k=1}\mathbb{E}\|e_t-\lambda e_{k,t}\|^2_2 \,.
\end{split}
\end{equation}
Replace \tcircle{1} and \tcircle{2} with their bounds and we have
\begin{equation}\label{eq:def step}
\begin{split}
    &\mathbb{E}[f(y_{t+1})-f(y_t)] \leq (-\frac{\eta}{2} + \eta^2 L)\mathbb{E}\|\nabla F_{\mathcal{S}}(y_t)\|^2_2 + (\frac{\eta L^2}{2} + \eta^2L^3)\frac{1}{K}\sum^{K}_{k=1}\mathbb{E}\|e_t-\lambda e_{k,t}\|^2_2 + \frac{\eta^2L\sigma^2}{2K} \\
    &\overset{(a)}{\leq} -\frac{\eta}{4}\mathbb{E}\|\nabla F_{\mathcal{S}}(y_t)\|^2_2 + \eta L^2 \cdot \frac{1}{K}\sum^{K}_{k=1}\mathbb{E}\|e_t-\lambda e_{k,t}\|^2_2 + \frac{\eta^2L\sigma^2}{2K} \,,
\end{split}
\end{equation}
where (a) is due to the assumption $\eta\leq \frac{1}{4L}$ for simplicity. Rearrange and sum from $t=0$ to $T-1$, we will have
\begin{equation}
\begin{split}
    &\frac{1}{T}\sum^{T-1}_{t=0}\mathbb{E}\|\nabla F_{\mathcal{S}}(y_t)\|^2_2 \leq \frac{4\mathbb{E}[F_{\mathcal{S}}(y_0)-F_{\mathcal{S}}(y_T)]}{\eta T} + \frac{2\eta L \sigma^2}{K} + 4L^2\cdot\frac{1}{KT}\sum^{T-1}_{t=0}\sum^{K}_{k=1}\mathbb{E}\|e_t-\lambda e_{k,t}\|^2_2 \\
    &\overset{(a)}{\leq} \frac{4\mathbb{E}[F_{\mathcal{S}}(y_0)-F_{\mathcal{S}}(y^*)]}{\eta T} + \frac{2\eta L \sigma^2}{K} + \frac{4}{(\sqrt{\frac{1-\delta/2}{1-\delta}}-1)^2}\eta^2L^2[B_1\sigma^2 + 2(1-\lambda)^2M^2] \\
    &= \frac{4\mathbb{E}[F_{\mathcal{S}}(y_0)-F_{\mathcal{S}}(y^*)]}{\eta T} + \frac{2\eta L \sigma^2}{K} + \frac{4}{(\sqrt{\frac{1-\delta/2}{1-\delta}}-1)^2}\eta^2L^2[\frac{K-1}{K^2}\sigma^2 + (\frac{1}{K}-\lambda)^2\sigma^2 + 2(1-\lambda)^2M^2] \,,
\end{split}
\end{equation}
where (a) follows Lemma \ref{lemma:error diff}. Let $\eta=\mathcal{O}(\sqrt{\frac{K}{T}})$, we have the convergence rate
\begin{equation}
    \frac{1}{T}\sum^{T-1}_{t=0}\mathbb{E}\|\nabla F_{\mathcal{S}}(y_t)\|^2_2 = \mathcal{O}(\frac{1}{\sqrt{KT}} + \frac{K}{T})\overset{K=\mathcal{O}(T^{1/3})}{=}\mathcal{O}(\frac{1}{\sqrt{KT}}) \,.
\end{equation}
If we are only interested in $\delta$, $\sigma^2$ and $M^2$, let $\lambda=\frac{\frac{1}{K}\sigma^2 + 2M^2}{\sigma^2 + 2M^2}$ and we have
\begin{equation}
    \frac{1}{T}\sum^{T-1}_{t=0}\mathbb{E}\|\nabla F_{\mathcal{S}}(y_t)\|^2_2 = \mathcal{O}(\frac{1}{(\sqrt{(1-\delta/2)/(1-\delta)}-1)^2}\cdot(\frac{K-1}{K^2}\sigma^2+\frac{2(1-\frac{1}{K})^2\sigma^2M^2}{\sigma^2 + 2M^2})) \,.
\end{equation}

\section{Proof of Generalization of DEF(-A) (Theorem \ref{def generalization})}\label{appendix:gen}
We use uniform stability to bound the generalization error. Let $\mathcal{S}=\{\xi_1,\xi_2,\cdots,\xi_N\}$ be the training dataset of size $N$, where each data $\xi_n$ is sampled from distribution $\mathcal{D}$. Let $\mathcal{S}^{(n)}=\{\xi^\prime_1,\xi^\prime_2,\cdots,\xi^\prime_N\}=\{\xi_1,\xi_2,\cdots,\xi_{n-1},\xi^\prime_{n},\xi_{n+1},\cdots,\xi_N\}$ be another training datasets of size $N$. We can see that $\mathcal{S}$ and $\mathcal{S}^{(n)}$ only differs in the $n^{\text{th}}$ data. Let the models trained on $\mathcal{S}$ and $\mathcal{S}^{(i)}$ be $x_t$ and $\widetilde{x}_t$ respectively for DEF-A. For DEF, they will be $y_t$ and $\widetilde{y}_t$ correspondingly.

In each iteration $t$ and under the same random sampling procedure, all workers select the same data from $\mathcal{S}$ and $\mathcal{S}^{(n)}$ with probability ${N-1 \choose K} / {N \choose K}=\frac{N-K}{N}$, while one of the workers selects different data from $\mathcal{S}$ and $\mathcal{S}^{(n)}$ with probability $1- {N-1 \choose K} / {N \choose K}=\frac{K}{N}$. For simplicity, let $G=\sqrt{\sigma^2+M^2}$, $B_2=|\frac{1}{K}-\lambda|+\frac{K-1}{K}$. Following \citet{hardt2016train} (Theorem 3.8), we only need to bound ($x_t$ for DEF-A and $y_t$ for DEF)
\begin{equation}\label{eq:hardt}
    \mathbb{E}[f(y_t;\xi)-f(\widetilde{y}_T;\xi)] \leq \frac{Kt_0}{N} + G\mathbb{E}[\|y_T-\widetilde{y}_T\|_2 | y_{t_0}-\widetilde{y}_{t_0}=0] \,.
\end{equation}

\subsection{Generalization Error of DEF}\label{sec:def gen}
In this section, we consider non-convex objective functions. We need Assumptions \ref{bounded variance}, \ref{bounded second moment}, \ref{lipschitz gradient}, \ref{approximate compressor}, \ref{allreduce compressor} and $\eta_t\leq\frac{c}{t+1}$. We first consider $y_t$ selecting the same data at iteration $t$.
\begin{equation}
\begin{split}
    &\|y_{t+1}-\widetilde{y}_{t+1}\|_2 = \|y_t-\eta_t g_t - \widetilde{y}_t + \eta_t\widetilde{g}_t\|_2 \leq \|y_t-\widetilde{y}_t\|_2 + \eta_t\|g_t-\widetilde{g}_t\|_2 \\
    &= \|y_t-\widetilde{y}_t\|_2 + \eta_t\|\frac{1}{K}\sum^{K}_{k=1}[\nabla f(x_t-\lambda e_{k,t};\xi_{k,t})-\nabla f(\widetilde{x}_t-\lambda \widetilde{e}_{k,t};\xi_{k,t})]\|_2 \\
    &\leq \|y_t-\widetilde{y}_t\|_2 + \frac{\eta_t}{K}\sum^{K}_{k=1}\|\nabla f(y_t+e_t-\lambda e_{k,t};\xi_{k,t})-\nabla f(\widetilde{y}_t+\widetilde{e}_t-\lambda \widetilde{e}_{k,t};\xi_{k,t})\|_2 \\
    &\leq (1+\eta_t L)\|y_t-\widetilde{y}_t\|_2 + \frac{\eta_tL}{K}\sum^{K}_{k=1}\|(e_t-\lambda e_{k,t})-(\widetilde{e}_t-\lambda \widetilde{e}_{k,t})\|_2 \\
    &= (1+\eta_t L)\|y_t-\widetilde{y}_t\|_2 + \frac{\eta_tL}{K}\sum^{K}_{k=1}\|(\frac{1}{K}-\lambda)(e_{k,t}-\widetilde{e}_{k,t}) + \frac{1}{K}\sum^{K}_{k^\prime=1,k^\prime\neq k}(e_{k^\prime,t}-\widetilde{e}_{k^\prime,t})\|_2 \\
    &\leq (1+\eta_t L)\|y_t-\widetilde{y}_t\|_2 + \frac{\eta_tLB_2}{K}\sum^{K}_{k=1} \underbrace{\|e_{k,t}-\widetilde{e}_{k,t}\|_2}_{\tcircle{1}} \,.
\end{split}
\end{equation}

Now we consider term \tcircle{1}. Following the same procedures in Lemma \ref{lemma:error bound}, but let $\eta_t\leq \frac{c}{t+1}$ and $\beta=\frac{\delta}{2(1-\delta)}$, we have
\begin{equation}
\begin{split}
    &\mathbb{E}\|e_{k,t}-\widetilde{e}_{k,t}\|^2_2 \leq (1-\delta)(1+\frac{1}{\beta})\cdot 2(\sigma^2+M^2)\sum^{t-1}_{t^\prime=0}[(1-\delta)(1+\beta)]^{t-1-t^\prime}\eta_{t^\prime}^2 \\
    &\leq 2(1-\delta)(1+\frac{1}{\beta})(\sigma^2+M^2)c^2\sum^{t-1}_{t^\prime=0}\frac{1}{(t^\prime+1)^2} \\
    &\leq 2(1-\delta)(1+\frac{1}{\beta})(\sigma^2+M^2)c^2[1+(-\frac{1}{t^\prime+1})|^{t-1}_{0}] \\
    &\leq 4(1-\delta)(1+\frac{1}{\beta})(\sigma^2+M^2)c^2 \\
    &= \frac{4(1-\delta)(2-\delta)}{\delta}G^2c^2 \,.
\end{split}
\end{equation}
Because $(\mathbb{E}\|e_{k,t}-\widetilde{e}_{k,t}\|_2)^2 \leq \mathbb{E}\|e_{k,t}-\widetilde{e}_{k,t}\|_2^2$, we have
\begin{equation}\label{eq:error diff bound}
\begin{split}
    \mathbb{E}\tcircle{1} \leq \sqrt{\mathbb{E}\|e_{k,t}-\widetilde{e}_{k,t}\|^2_2} \leq 2Gc\underbrace{\sqrt{\frac{(1-\delta)(2-\delta)}{\delta}}}_{B_4} = 2B_4Gc\,.
\end{split}
\end{equation}
At iteration $t$, if a worker $k^\prime\in[K]$ selects different data from $\mathcal{S}$ and $\mathcal{S}^{(n)}$, we have
\begin{equation}
\begin{split}
    &\|y_{t+1}-\widetilde{y}_{t+1}\|_2 \leq \|y_t-\widetilde{y}_t\|_2 + \eta_t\|g_t-\widetilde{g}_t\|_2 \leq \|y_t-\widetilde{y}_t\|_2 + 2G\eta_t \,.
\end{split}
\end{equation}
When we consider both circumstances, we have
\begin{equation}
\begin{split}
    &\mathbb{E}\|y_{t+1}-\widetilde{y}_{t+1}\|_2 \leq (1-\frac{K}{N})[(1+\eta_t L)\mathbb{E}\|y_t-\widetilde{y}_t\|_2 + \eta_tLB_2 \cdot 2B_4Gc] + \frac{K}{N}[\mathbb{E}\|y_t-\widetilde{y}_t\|_2 + 2G\eta_t] \\
    &= [1+(1-\frac{K}{N})\eta_t L]\mathbb{E}\|y_t-\widetilde{y}_t\|_2 + \underbrace{[\frac{2K}{N}G + 2(1-\frac{K}{N})LB_2B_4Gc]}_{B_5}\eta_t \\
    &\overset{(a)}{\leq} \exp((1-\frac{K}{N})\eta_tL)\mathbb{E}\|y_t-\widetilde{y}_t\|_2 + B_5\eta_t \,.
\end{split}
\end{equation}
Unwind the recurrence with $t=0,1,\cdots,T-1$, we have
\begin{equation}
\begin{split}
    &\mathbb{E}\|y_T-\widetilde{y}_T\|_2 \leq \sum^{T-1}_{t=t_0}B_5\frac{c}{t+1}\prod^{T-1}_{t^\prime=t+1}\exp((1-\frac{K}{N})\frac{c}{t+1}L) \\
    &= B_5c\sum^{T-1}_{t=t_0}\frac{1}{t+1}\exp((1-\frac{K}{N})Lc\sum^{T-1}_{t^\prime=t+1}\frac{1}{t+1}) \\
    &\overset{(a)}{\leq} B_5c\sum^{T-1}_{t=t_0}\frac{1}{t+1}\exp((1-\frac{K}{N})Lc\log\frac{T}{t+1}) \\
    &=B_5cT^{(1-\frac{K}{N})Lc}\sum^{T-1}_{t=t_0}(t+1)^{-(1-\frac{K}{N})Lc-1} \\
    &= B_5cT^{(1-\frac{K}{N})Lc} \frac{t_0^{-(1-\frac{K}{N})Lc}-T^{-(1-\frac{K}{N})Lc}}{(1-\frac{K}{N})Lc} \\
    &\leq \frac{B_5}{(1-\frac{K}{N})L}(\frac{T}{t_0})^{(1-\frac{K}{N})Lc}
\end{split}
\end{equation}
where (a) is due to $\sum^{T-1}_{t^{\prime}=t+1}\frac{1}{t^{\prime}+1}\leq \int^{T-1}_{t}\log (t^{\prime}+1)dt^{\prime}$. Following Eq.~(\ref{eq:hardt}),
\begin{equation}
    \mathbb{E}[f(y_t;\xi)-f(\widetilde{y}_T;\xi)] \leq \frac{Kt_0}{N} + \underbrace{\frac{B_5G}{(1-\frac{K}{N})L}}_{B_6}(\frac{T}{t_0})^{(1-\frac{K}{N})Lc} \,.
\end{equation}
The R.H.S is minimized when
\begin{equation}
    t_0=[(\frac{N}{K}-1)LcB_6]^{1/((1-\frac{K}{N})Lc+1)} T^{(1-\frac{K}{N})Lc/((1-\frac{K}{N})Lc+1)} \,,
\end{equation}
which gives us
\begin{equation}
\begin{split}
    \mathbb{E}[f(y_T;\xi)-f(\widetilde{y}_T;\xi)] \leq [\frac{K}{N}+\frac{1}{(\frac{N}{K}-1)Lc}][(\frac{N}{K}-1)LcB_6]^{1/((1-\frac{K}{N})Lc+1)} T^{(1-\frac{K}{N})Lc/((1-\frac{K}{N})Lc+1)} \,.
\end{split}
\end{equation}
Note that when $K=1$, $\lambda=1$, we will have $B_2=0$, $B_5=\frac{2K}{N}G$, and $B_6=\frac{2G^2}{(N-1)L}$. Then the R.H.S. equals
\begin{equation}
\begin{split}
    &[\frac{1}{N}+\frac{1}{(N-1)Lc}](2cG^2)^{1/((1-\frac{1}{N})Lc+1)} T^{(1-\frac{1}{N})Lc/((1-\frac{1}{N})Lc+1)} \\
    &= \frac{1+1/(Lc)}{N-1} T (\frac{2cG^2}{T})^{1/((1-\frac{1}{N})Lc+1)} \\
    &\overset{(a)}{\leq} \frac{1+1/(Lc)}{N-1} T (\frac{2cG^2}{T})^{1/(Lc+1)} \\
    &= \frac{1+1/(Lc)}{N-1}(2cG^2)^{1/(Lc+1)}T^{Lc/(Lc+1)} \,,
\end{split}
\end{equation}
which matches the result in \citet{hardt2016train} for SGD. (a) is due to $\frac{2cG^2}{T}\leq 1$ when $t_0\leq T$ .

\subsection{Generalization Error of DEF-A}\label{sec:defa gen}
In this section, we consider non-convex objective functions and random sparsification which satisfies Assumptions \ref{approximate compressor} and \ref{allreduce compressor}. We need Assumptions \ref{bounded variance}, \ref{bounded second moment}, \ref{lipschitz gradient}, and $\eta_t\leq\frac{c}{t+1}$.

Now we consider $x_t$.
\begin{equation}\label{eq:defa gen bound 1}
\begin{split}
    &\mathbb{E}\|x_{t+1}-\widetilde{x}_{t+1}\|_2 \overset{(a)}{=} \mathbb{E}\|x_t-\mathcal{C}(\eta_t g_t+e_t) - \widetilde{x}_t+\mathcal{C}(\eta_t \widetilde{g}_t + \widetilde{e}_t)\|_2\\
    &\overset{(a)}{\leq} \mathbb{E}\|x_t-\widetilde{x}_t\|_2 + \mathbb{E}\|\mathcal{C}(\eta_tg_t-\eta_t\widetilde{g}_t)\|_2 + \mathbb{E}\|\mathcal{C}(e_t-\widetilde{e}_t)\|_2 \\
    &\leq \mathbb{E}\|x_t-\widetilde{x}_t\|_2 + \mathbb{E}\sqrt{\mathbb{E}_{\mathcal{C}}\|\mathcal{C}(\eta_tg_t-\eta_t\widetilde{g}_t)\|^2_2} + \mathbb{E}\sqrt{\mathbb{E}_{\mathcal{C}}\|\mathcal{C}(e_t-\widetilde{e}_t)\|^2_2}\\
    &\overset{(a)}{=} \mathbb{E}\|x_t-\widetilde{x}_t\|_2 + \mathbb{E}\sqrt{\delta\eta_t^2\|g_t-\widetilde{g}_t\|^2_2} + \mathbb{E}\sqrt{\delta\|e_t-\widetilde{e}_t\|^2_2} \\
    &= \mathbb{E}\|x_t-\widetilde{x}_t\|_2 + \delta^{\frac{1}{2}}\eta_t\mathbb{E}\|g_t-\widetilde{g}_t\|_2 + \delta^{\frac{1}{2}}\mathbb{E}\|e_t-\widetilde{e}_t\|_2 \\
    &\leq \mathbb{E}\|x_t-\widetilde{x}_t\|_2 + \delta^{\frac{1}{2}}\eta_t\mathbb{E}\|g_t-\widetilde{g}_t\|_2 + \delta^{\frac{1}{2}}\frac{1}{K}\sum^{K}_{k=1}\mathbb{E}\|e_{k,t}-\widetilde{e}_{k,t}\|_2 \,.
\end{split}
\end{equation}
where (a) is due to the random sparsification. When selecting the same data at iteration $t$, we have
\begin{equation}
\begin{split}
    &\mathbb{E}\|x_{t+1}-\widetilde{x}_{t+1}\|_2\\
    &\leq \mathbb{E}\|x_t-\widetilde{x}_t\|_2 + \frac{\delta^{\frac{1}{2}}}{K}\sum^{K}_{k=1}\mathbb{E}\|e_{k,t}-\widetilde{e}_{k,t}\|_2 + \delta^{\frac{1}{2}}\eta_t\mathbb{E}\|\frac{1}{K}\sum^{K}_{k=1}[\nabla f(x_t-\lambda e_{k,t};\xi_{k,t})-\nabla f(\widetilde{x}_t-\lambda \widetilde{e}_{k,t};\xi_{k,t})]\|_2 \\
    &\leq \mathbb{E}\|x_t-\widetilde{x}_t\|_2 + \frac{\delta^{\frac{1}{2}}}{K}\sum^{K}_{k=1}\mathbb{E}\|e_{k,t}-\widetilde{e}_{k,t}\|_2 + \frac{\delta^{\frac{1}{2}}\eta_tL}{K}\sum^{K}_{k=1}\|x_t-\lambda e_{k,t}-\widetilde{x}_t+\lambda \widetilde{e}_{k,t}\|_2 \\
    &\leq (1+\delta^{\frac{1}{2}}\eta_tL)\mathbb{E}\|x_t-\widetilde{x}_t\|_2 + \frac{\delta^{\frac{1}{2}}(1+\eta_t L\lambda)}{K}\sum^{K}_{k=1}\mathbb{E}\|e_{k,t}-\widetilde{e}_{k,t}\|_2 \,.
\end{split}
\end{equation}
When selecting different data at iteration $t$, we have
\begin{equation}
\begin{split}
    \mathbb{E}\|x_{t+1}-\widetilde{x}_{t+1}\|_2 \leq \mathbb{E}\|x_t-\widetilde{x}_t\|_2 + \delta^{\frac{1}{2}}\eta_t\cdot2G + \frac{\delta^{\frac{1}{2}}}{K}\sum^{K}_{k=1}\mathbb{E}\|e_{k,t}-\widetilde{e}_{k,t}\|_2 \,.
\end{split}
\end{equation}
When we consider both circumstances, we have
\begin{equation}
\begin{split}
    &\mathbb{E}\|x_{t+1}-\widetilde{x}_{t+1}\|_2 \leq (1-\frac{K}{N})[(1+\delta^{\frac{1}{2}}\eta_tL)\mathbb{E}\|x_t-\widetilde{x}_t\|_2 + \frac{\delta^{\frac{1}{2}}(1+\eta_t L\lambda)}{K}\sum^{K}_{k=1}\mathbb{E}\|e_{k,t}-\widetilde{e}_{k,t}\|_2] \\
    &\quad\quad\quad\quad\quad\quad\quad\quad+ \frac{K}{N}[\mathbb{E}\|x_t-\widetilde{x}_t\|_2 + \delta^{\frac{1}{2}}\eta_t\cdot2G+ \frac{\delta^{\frac{1}{2}}}{K}\sum^{K}_{k=1}\mathbb{E}\|e_{k,t}-\widetilde{e}_{k,t}\|_2] \\
    &=[1+(1-\frac{K}{N})\delta^{\frac{1}{2}}\eta_tL]\mathbb{E}\|x_t-\widetilde{x}_t\|_2 + \frac{K}{N}2\delta^{\frac{1}{2}}\eta_tG + \delta^{\frac{1}{2}}[1+(1-\frac{K}{N})\eta_tL\lambda]\frac{1}{K}\sum^{K}_{k=1}\mathbb{E}\|e_{k,t}-\widetilde{e}_{k,t}\|_2 \\
    &\overset{(a)}{\leq} [1+(1-\frac{K}{N})\delta^{\frac{1}{2}}\eta_tL]\mathbb{E}\|x_t-\widetilde{x}_t\|_2 + \frac{K}{N}2\delta^{\frac{1}{2}}\eta_tG + \delta^{\frac{1}{2}}[1+(1-\frac{K}{N})\eta_tL\lambda] \cdot 2B_4Gc \\
    &\overset{(b)}{\leq}\exp((1-\frac{K}{N})\delta^{\frac{1}{2}}\eta_t L)\mathbb{E}\|x_t-\widetilde{x}_t\|_2 + \frac{K}{N}2\delta^{\frac{1}{2}}\eta_tG + \delta^{\frac{1}{2}}[1+(1-\frac{K}{N})\eta_tL\lambda] \cdot 2B_4Gc \,,
\end{split}
\end{equation}
where (a) follows the Eq.~(\ref{eq:error diff bound}). Let $\eta_t\leq\frac{c}{t+1}$, we have
\begin{equation}
\begin{split}
    &\mathbb{E}\|x_T-\widetilde{x}_T\|_2 \leq \sum^{T-1}_{t=t_0}[\frac{K}{N}2\delta^{\frac{1}{2}}\eta_tG + \delta^{\frac{1}{2}}(1+(1-\frac{K}{N})\eta_tL\lambda)\cdot2B_4Gc]\prod^{T-1}_{t^{\prime}=t+1}\exp((1-\frac{K}{N})\delta^{\frac{1}{2}}\eta_{t^{\prime}}L) \\
    &\leq 2\delta^{\frac{1}{2}} G\sum^{T-1}_{t=t_0}[\frac{K}{N}\frac{c}{t+1} + (1+(1-\frac{K}{N})\frac{c}{t+1}L\lambda)B_4c] \exp((1-\frac{K}{N})\delta^{\frac{1}{2}} Lc\sum^{T-1}_{t^{\prime\prime}=t+1}\frac{1}{t^{\prime\prime}+1}) \\
    &\leq 2\delta^{\frac{1}{2}} G\sum^{T-1}_{t=t_0}[\frac{K}{N}\frac{c}{t+1} + (1+(1-\frac{K}{N})\frac{c}{t+1}L\lambda)B_4c] \exp((1-\frac{K}{N})\delta^{\frac{1}{2}} Lc\log(\frac{T}{t+1})) \\
    &= 2\delta^{\frac{1}{2}} G\sum^{T-1}_{t=t_0}[\frac{K}{N}\frac{c}{t+1} + (1+(1-\frac{K}{N})\frac{c}{t+1}L\lambda)B_4c] (\frac{T}{t+1})^{(1-\frac{K}{N})\delta^{\frac{1}{2}} Lc} \\
    &\leq 2\delta^{\frac{1}{2}} Gc[\frac{K}{N}+1+(1-\frac{K}{N})Lc\lambda B_4]T^{(1-\frac{K}{N})\delta^{\frac{1}{2}} Lc}\sum^{T-1}_{t=t_0}(t+1)^{-(1-\frac{K}{N})\delta^{\frac{1}{2}} Lc} \\
    &\leq 2\delta^{\frac{1}{2}} Gc[\frac{K}{N}+1+(1-\frac{K}{N})Lc\lambda B_4]T^{(1-\frac{K}{N})\delta^{\frac{1}{2}} Lc} \frac{t_0^{-1-(1-\frac{K}{N})\delta^{\frac{1}{2}} Lc}-T^{-1-(1-\frac{K}{N})\delta^{\frac{1}{2}} Lc}}{1+(1-\frac{K}{N})\delta^{\frac{1}{2}} Lc} \\
    &\leq \frac{2\delta^{\frac{1}{2}} G[\frac{K}{N}+1+(1-\frac{K}{N})Lc\lambda B_4]}{1+(1-\frac{K}{N})\delta^{\frac{1}{2}} L} (\frac{T}{t_0})^{(1-\frac{K}{N})\delta^{\frac{1}{2}} Lc} \,.
\end{split}
\end{equation}
Following Eq.~(\ref{eq:hardt}),
\begin{equation}
\begin{split}
    \mathbb{E}[f(x_T;\xi)-f(\widetilde{x}_T;\xi)] \leq \frac{Kt_0}{N} + \underbrace{\frac{2\delta^{\frac{1}{2}} G[\frac{K}{N}+1+(1-\frac{K}{N})Lc\lambda B_4]}{1+(1-\frac{K}{N})\delta^{\frac{1}{2}} L}}_{B_7} (\frac{T}{t_0})^{(1-\frac{K}{N})\delta^{\frac{1}{2}} Lc} \,.
\end{split}
\end{equation}
The R.H.S is minimized when
\begin{equation}
    t_0=[(\frac{N}{K}-1)LcB_7]^{1/((1-\frac{K}{N})\delta^{\frac{1}{2}} Lc+1)} T^{(1-\frac{K}{N})\delta^{\frac{1}{2}} Lc/((1-\frac{K}{N})\delta^{\frac{1}{2}} Lc+1)} \,,
\end{equation}
which gives us
\begin{equation}
\begin{split}
    \mathbb{E}[f(x_T;\xi)-f(\widetilde{x}_T;\xi)] \leq [\frac{K}{N}+\frac{1}{(\frac{N}{K}-1)Lc}][(\frac{N}{K}-1)LcB_7]^{1/((1-\frac{K}{N})\delta^{\frac{1}{2}} Lc+1)} T^{(1-\frac{K}{N})\delta^{\frac{1}{2}} Lc/((1-\frac{K}{N})\delta^{\frac{1}{2}} Lc+1)} \,.
\end{split}
\end{equation}

\subsection{Optimization Error of DEF}\label{sec:def opt}
In this section, we consider non-convex objective functions satisfying the Polyak-\L{}ojasiewicz (PL) condition, which establishes the relation between the objective function and the gradient norm. We consider $\delta$-approximate and ring-allreduce compatible compressor. We need Assumptions \ref{bounded variance}, \ref{bounded second moment}, \ref{lipschitz gradient}, \ref{approximate compressor}, \ref{allreduce compressor}, and $\eta_t=\frac{c}{t+1}\leq\frac{1}{4L}$. From Eq.~(\ref{eq:def step}) and the PL condition, we have
\begin{equation}
\begin{split}
    &\mathbb{E}[F_{\mathcal{S}}(y_{t+1})-F_{\mathcal{S}}(y_t)] \leq -\frac{\eta_t}{4}\mathbb{E}\|\nabla F_{\mathcal{S}}(y_t)\|^2_2 + \eta_t L^2 \cdot \frac{1}{K}\sum^{K}_{k=1}\mathbb{E}\|e_t-\lambda e_{k,t}\|^2_2 + \frac{\eta_t^2L\sigma^2}{2K} \\
    &\leq -\frac{\mu\eta_t}{2}\mathbb{E}[F_{\mathcal{S}}(y_t)-F_{\mathcal{S}}(y^*)] + \eta_t L^2 \cdot \frac{1}{K}\sum^{K}_{k=1}\mathbb{E}\|e_t-\lambda e_{k,t}\|^2_2 + \frac{\eta_t^2L\sigma^2}{2K} \,,
\end{split}
\end{equation}
where we need $\eta_t\leq \frac{1}{4L}$ following Section \ref{appendix:convergence}. Rearrange,
\begin{equation}
\begin{split}
    &\mathbb{E}[F_{\mathcal{S}}(y_{t+1})-F_{\mathcal{S}}(y^*)] \leq (1-\frac{\mu\eta_t}{2})\mathbb{E}[F_{\mathcal{S}}(y_t)-F_{\mathcal{S}}(y^*)] + \eta_t L^2 \cdot \frac{1}{K}\sum^{K}_{k=1}\mathbb{E}\|e_t-\lambda e_{k,t}\|^2_2 + \frac{\eta_t^2L\sigma^2}{2K} \\
    &\overset{(a)}{\leq} (1-\frac{\mu\eta_t}{2})\mathbb{E}[F_{\mathcal{S}}(y_t)-F_{\mathcal{S}}(y^*)] + \frac{B_1\sigma^2 + 2(1-\lambda)^2M^2}{(\sqrt{\frac{1-\delta/2}{1-\delta}}-1)^2}\eta_t^3L^2 + \frac{\eta_t^2L\sigma^2}{2K} \,,
\end{split}
\end{equation}
where (a) follows Lemma \ref{lemma:error diff}. Let $\eta_t=\frac{c}{t+1}$, we have
\begin{equation}
\begin{split}
    &\mathbb{E}[F_{\mathcal{S}}(y_{T})-F_{\mathcal{S}}(y^*)]\\
    &\leq \mathbb{E}[F_{\mathcal{S}}(y_t)-F_{\mathcal{S}}(y^*)]\underbrace{\prod^{T-1}_{t=0}(1-\frac{\mu c}{2(t+1)})}_{\tcircle{1}} + [\frac{B_1\sigma^2 + 2(1-\lambda)^2M^2}{(\sqrt{\frac{1-\delta/2}{1-\delta}}-1)^2} + \frac{\sigma^2}{2K}]L\underbrace{\sum^{T-1}_{t=0}\frac{c^2}{(t+1)^2}\prod^{T-1}_{t^\prime=t+1}(1-\frac{\mu c}{2(t+1)})}_{\tcircle{2}} \,,
\end{split}
\end{equation}
where
\begin{equation}
\begin{split}
    \tcircle{1} \leq \prod^{T-1}_{t=0}\exp(-\frac{\mu c}{2(t+1)}) = \exp(-\frac{\mu c}{2})\exp(-\frac{\mu c}{2}\sum^{T-1}_{t=1}\frac{1}{t+1}) \leq \exp(-\frac{\mu c}{2})\exp(-\frac{\mu c}{2}\log T) = \exp(-\frac{\mu c}{2})T^{-\frac{\mu c}{2}} \,,
\end{split}
\end{equation}
\begin{equation}
\begin{split}
    &\tcircle{2} \leq \sum^{T-1}_{t=0} \frac{c^2}{(t+1)^2} \exp(-\frac{\mu c}{2}\sum^{T-1}_{t^\prime=t+1}\frac{1}{t^\prime+1}) \leq \sum^{T-1}_{t=0} \frac{c^2}{(t+1)^2} \exp(-\frac{\mu c}{2} \log\frac{T}{t+1}) = c^2T^{-\frac{\mu c}{2}}\sum^{T-1}_{t=0}(t+1)^{\frac{\mu c}{2}-2} \,.
\end{split}
\end{equation}
When $\frac{\mu c}{2}-2\geq 0$,
\begin{equation}
\begin{split}
    \tcircle{2} \leq \frac{c^2T^{-\frac{\mu c}{2}}}{\frac{\mu c}{2}-1}((T+1)^{\frac{\mu c}{2}-1}-1) \leq \frac{c^2}{\frac{\mu c}{2}-1}(\frac{T+1}{T})^{\frac{\mu c}{2}-1}T^{-1} \,.
\end{split}
\end{equation}
When $\frac{\mu c}{2}-2\leq 0$ and $\frac{\mu c}{2}-2 \neq -1$,
\begin{equation}
\begin{split}
    \tcircle{2} \leq \frac{c^2T^{-\frac{\mu c}{2}}}{\frac{\mu c}{2}-1}(1+T^{\frac{\mu c}{2}-1}-1) = \frac{c^2}{\frac{\mu c}{2}-1}T^{-1} \,.
\end{split}
\end{equation}
When $\frac{\mu c}{2}-2=-1$,
\begin{equation}
\begin{split}
    \tcircle{2} \leq c^2T^{-1}(1+\log T) \,.
\end{split}
\end{equation}
Hence,
\begin{equation}
\begin{split}
    \mathbb{E}[F_{\mathcal{S}}(y_T)-F_{\mathcal{S}}(y^*)]=\widetilde{\mathcal{O}}(T^{-\frac{\mu c}{2}}+T^{-1}) \,.
\end{split}
\end{equation}

\subsection{Optimization Error of DEF-A}\label{sec:defa opt}
In this section, we consider non-convex objective functions satisfying the Polyak-\L{}ojasiewicz (PL) condition, which establishes the relation between the objective function and the gradient norm. The compressor we consider here is random sparsification which satisfies Assumptions \ref{approximate compressor} and \ref{allreduce compressor}. We need Assumptions \ref{bounded variance}, \ref{bounded second moment}, \ref{lipschitz gradient}, and $\eta_t=\frac{c}{t+1}\leq \frac{1}{8L}$.
\begin{equation}
\begin{split}
    &\mathbb{E}[F_{\mathcal{S}}(x_{t+1})-F_{\mathcal{S}}(x_t)] \leq \mathbb{E}\langle\nabla F_{\mathcal{S}}(x_t),x_{t+1}-x_t\rangle + \frac{L}{2}\mathbb{E}\|x_{t+1}-x_t\|^2_2 \\
    &\overset{(a)}{=} \underbrace{\mathbb{E}\langle\nabla F_{\mathcal{S}}(x_t),-\mathcal{C}(\eta_t g_t+e_t)\rangle}_{\tcircle{1}} + \frac{L}{2}\underbrace{\mathbb{E}\|\mathcal{C}(\eta_t g_t +e_t)\|^2_2}_{\tcircle{2}} \,,
\end{split}
\end{equation}
where (a) follows Assumption \ref{allreduce compressor}. For term \tcircle{1},
\begin{equation}\label{eq:defa opt 1}
\begin{split}
    &\tcircle{1} \overset{(a)}{=} -\mathbb{E}\langle\nabla F_{\mathcal{S}}(x_t),\delta(\eta_t g_t + e_t)\rangle = -\delta\eta_t\mathbb{E}\langle\nabla F_{\mathcal{S}}(x_t), \frac{1}{K}\sum^{K}_{k=1}\nabla F_{\mathcal{S}}(x_t-\lambda e_{k,t})+\frac{e_t}{\eta_t}\rangle \\
    &= -\delta\eta_t\mathbb{E}\|\nabla F_{\mathcal{S}}(x_t)\|^2_2 - \delta\eta_t\mathbb{E}\langle\nabla F_{\mathcal{S}}(x_t), \frac{1}{K}\sum^{K}_{k=1}\nabla F_{\mathcal{S}}(x_t-\lambda e_{k,t})+\frac{e_t}{\eta_t}-\nabla F_{\mathcal{S}}(x_t)\rangle \\
    &\leq -\frac{\delta\eta_t}{2}\mathbb{E}\|\nabla F_{\mathcal{S}}(x_t)\|^2_2 + \frac{\delta\eta_t}{2}\mathbb{E}\|\frac{1}{K}\sum^{K}_{k=1}\nabla F_{\mathcal{S}}(x_t-\lambda e_{k,t})-\nabla F_{\mathcal{S}}(x_t) + \frac{e_t}{\eta_t}\|^2_2 \\
    &\leq -\frac{\delta\eta_t}{2}\mathbb{E}\|\nabla F_{\mathcal{S}}(x_t)\|^2_2 + \frac{\delta\eta_t L^2}{2K}\sum^{K}_{k=1}\mathbb{E}\|\lambda e_{k,t}\|^2_2 + \frac{\delta}{2\eta_t}\mathbb{E}\|e_t\|^2_2 \\
    &\leq -\frac{\delta\eta_t}{2}\mathbb{E}\|\nabla F_{\mathcal{S}}(x_t)\|^2_2 + \frac{\delta(\eta_t^2L^2\lambda^2+1)}{2\eta_t K}\sum^{K}_{k=1}\mathbb{E}\|e_{k,t}\|^2_2\,,
\end{split}
\end{equation}
where (a) is due to the random sparsification compressor. For term \tcircle{2},
\begin{equation}\label{eq:defa opt 2}
\begin{split}
    &\tcircle{2} \overset{(a)}{=} \delta\mathbb{E}\|\eta_t g_t + e_t\|^2_2 \leq 2\delta\eta_t^2\mathbb{E}\|g_t\|^2_2 + 2\delta \mathbb{E}\|e_t\|^2_2\\
    &= 2\delta\eta_t^2\mathbb{E}\|\frac{1}{K}\sum^{K}_{k=1}\nabla F_{\mathcal{S}}(x_t-\lambda e_{k,t})\|^2_2 + \frac{2\delta\eta_t^2\sigma^2}{K} + 2\delta\mathbb{E}\|e_t\|^2_2 \\
    &\leq \frac{2\delta\eta_t^2}{K}\sum^{K}_{k=1}\mathbb{E}\|\nabla F_{\mathcal{S}}(x_t-\lambda e_{k,t})-\nabla F_{\mathcal{S}}(x_t) + \nabla F_{\mathcal{S}}(x_t)\|^2_2 + 2\delta\mathbb{E}\|e_t\|^2_2 + \frac{2\delta\eta_t^2\sigma^2}{K} \\
    &\leq 4\delta\eta_t^2\mathbb{E}\|\nabla F_{\mathcal{S}}(x_t)\|^2_2 + \frac{4\delta\eta_t^2L^2}{K}\sum^{K}_{k=1}\mathbb{E}\|\lambda e_{k,t}\|^2_2 + 2\delta\mathbb{E}\|e_t\|^2_2 + \frac{2\delta\eta_t^2\sigma^2}{K} \\
    &= 4\delta\eta_t^2\mathbb{E}\|\nabla F_{\mathcal{S}}(x_t)\|^2_2 + \frac{2\delta(2\eta_t^2L^2\lambda^2+1)}{K}\sum^{K}_{k=1}\mathbb{E}\|e_{k,t}\|^2_2 + \frac{2\delta\eta_t^2\sigma^2}{K} \,,
\end{split}
\end{equation}
where (a) is due to the random sparsification compressor. Put them together, let $\eta_t \leq \frac{1}{8L}$, and we have
\begin{equation}
\begin{split}
    &\mathbb{E}[F_{\mathcal{S}}(x_{t+1})-F_{\mathcal{S}}(x_t)]\\
    &\leq -\frac{\delta\eta_t}{2}(1-4\eta_t L)\mathbb{E}\|\nabla F_{\mathcal{S}}(x_t)\|^2_2 + \frac{\delta(1+2\eta_t L)(1+2\eta_t^2L^2\lambda^2)}{2\eta_t K}\sum^{K}_{k=1}\mathbb{E}\|e_{k,t}\|^2_2 + \frac{\delta\eta_t^2L\sigma^2}{K} \\
    &\leq -\frac{\delta\eta_t}{4}\mathbb{E}\|\nabla F_{\mathcal{S}}(x_t)\|^2_2 + \frac{\delta(1+\lambda^2)}{\eta_t }\cdot\frac{1}{K}\sum^{K}_{k=1}\mathbb{E}\|e_{k,t}\|^2_2 + \frac{\delta\eta_t^2L\sigma^2}{K} \\
    &\overset{(a)}{\leq} -\frac{\mu\delta\eta_t}{2}\mathbb{E}[F_{\mathcal{S}}(x_t)-F_{\mathcal{S}}(x^*)] + \frac{\delta(1+\lambda^2)\eta_t(\sigma^2+M^2)}{(1/\sqrt{1-\delta}-1)^2} + \frac{\delta\eta_t^2L\sigma^2}{K} \,,
\end{split}
\end{equation}
where (a) is due to $\|\nabla F_{\mathcal{S}}(x_t)\|^2_2\geq 2\mu(F_{\mathcal{S}}(x_t)-F_{\mathcal{S}}(x^*))$ according to the PL condition and Lemma \ref{lemma:error bound}. Rearrange,
\begin{equation}
\begin{split}
    &\mathbb{E}[F_{\mathcal{S}}(x_{t+1})-F_{\mathcal{S}}(x^*)] \leq (1-\frac{\mu\delta\eta_t}{2})\mathbb{E}[F_{\mathcal{S}}(x_t)-F_{\mathcal{S}}(x^*)] + \frac{\delta(1+\lambda^2)\eta_t(\sigma^2+M^2)}{(1/\sqrt{1-\delta}-1)^2} + \frac{\delta\eta_t^2L\sigma^2}{K} \,. \\
\end{split}
\end{equation}
Let $\eta_t=\frac{c}{t+1}$ and $G=\sqrt{\sigma^2+M^2}$. Take this recurrence from $t=0$ to $T-1$ and we have
\begin{equation}
\begin{split}
    &\mathbb{E}[F_{\mathcal{S}}(x_T)-F_{\mathcal{S}}(x^*)] \leq \mathbb{E}[F_{\mathcal{S}}(x_0)-F_{\mathcal{S}}(x^*)]\underbrace{\prod^{T-1}_{t=0}(1-\frac{\mu\delta c}{2(t+1)})}_{\tcircle{3}} \\
    & + \frac{\delta(1+\lambda^2)G^2}{(1/\sqrt{1-\delta}-1)^2}\underbrace{\sum^{T-1}_{t^\prime=0}\frac{c}{t^\prime+1} \underbrace{\prod^{T-1}_{t=t^\prime+1}(1-\frac{\mu\delta c}{2(t+1)})}_{\tcircle{4}}}_{\tcircle{5}} + \frac{\delta L\sigma^2}{K}\underbrace{\sum^{T-1}_{t^\prime=0}\frac{c^2}{(t^\prime+1)^2} \underbrace{\prod^{T-1}_{t=t^\prime+1}(1-\frac{\mu\delta c}{2(t+1)})}_{\tcircle{4}}}_{\tcircle{6}} \,,
\end{split}
\end{equation}
where
\begin{equation}
\begin{split}
    \tcircle{3} &\leq \prod^{T-1}_{t=0}\exp(-\frac{\mu\delta c}{2(t+1)}) = \exp(-\frac{\mu\delta c}{2}\sum^{T-1}_{t=0}\frac{1}{t+1}) \leq \exp(-\frac{\mu\delta c}{2})\exp(-\frac{\mu\delta c}{2}\sum^{T-1}_{t=1}\frac{1}{t+1}) \\
    &\leq \exp(-\frac{\mu\delta c}{2})\exp(-\frac{\mu\delta c}{2}\log(T)) = \exp(-\frac{\mu\delta c}{2})T^{-\frac{\mu\delta c}{2}} \,.
\end{split}
\end{equation}
\begin{equation}
\begin{split}
    \tcircle{4} &\leq \prod^{T-1}_{t=t^\prime+1}\exp(-\frac{\mu\delta c}{2(t+1)}) = \exp(-\frac{\mu\delta c}{2}\sum^{T-1}_{t=t^\prime+1}\frac{1}{t+1}) \leq \exp(-\frac{\mu\delta c}{2}\log(\frac{T}{t^\prime+1}))= (\frac{T}{t^\prime+1})^{-\frac{\mu\delta c}{2}} \,,
\end{split}
\end{equation}
When $\frac{\mu\delta c}{2}\geq 1$,
\begin{equation}
\begin{split}
    \tcircle{5} \leq cT^{-\frac{\mu\delta c}{2}}\sum^{T-1}_{t^\prime=0}(t^\prime+1)^{\frac{\mu\delta c}{2}-1} \leq cT^{-\frac{\mu\delta c}{2}} \cdot \frac{2}{\mu\delta c}[(T+1)^{\frac{\mu\delta c}{2}}-1] \leq \frac{2}{\mu\delta}(\frac{T+1}{T})^{\frac{\mu\delta c}{2}} \,.
\end{split}
\end{equation}
When $0 < \frac{\mu\delta c}{2} \leq 1$,
\begin{equation}
\begin{split}
    \tcircle{5} \leq cT^{-\frac{\mu\delta c}{2}}\sum^{T-1}_{t^\prime=0}(t^\prime+1)^{\frac{\mu\delta c}{2}-1} \leq cT^{-\frac{\mu\delta c}{2}} \cdot [1 + \frac{2}{\mu\delta c}(T^{\frac{\mu\delta c}{2}}-1)] \leq \frac{2}{\mu\delta} \,.
\end{split}
\end{equation}
When $\frac{\mu\delta c}{2} \geq 2$,
\begin{equation}
\begin{split}
    \tcircle{6} \leq c^2T^{-\frac{\mu\delta c}{2}}\sum^{T-1}_{t^\prime=0}(t^\prime+1)^{\frac{\mu\delta c}{2}-2} \leq c^2T^{-\frac{\mu\delta c}{2}} \cdot \frac{1}{\mu\delta c/2-1}[(T+1)^{\frac{\mu\delta c}{2}-1}-1] \leq \frac{c^2}{\mu\delta c/2-1} (\frac{T+1}{T})^{\frac{\mu\delta c}{2}-1} T^{-1} \,.
\end{split}
\end{equation}
When $\frac{\mu\delta c}{2}\leq 2$, $\frac{\mu\delta c}{2}\neq 1$,
\begin{equation}
\begin{split}
    \tcircle{6} \leq c^2T^{-\frac{\mu\delta c}{2}}\sum^{T-1}_{t^\prime=0}(t^\prime+1)^{\frac{\mu\delta c}{2}-2} \leq c^2T^{-\frac{\mu\delta c}{2}} \cdot [1 + \frac{1}{\mu\delta c/2-1}(T^{\frac{\mu\delta c}{2}-1}-1)] \leq \frac{c^2}{\mu\delta c/2-1}T^{-1} \,.
\end{split}
\end{equation}
When $\frac{\mu\delta c}{2}=1$,
\begin{equation}
\begin{split}
    \tcircle{6} \leq c^2T^{-1}\sum^{T-1}_{t^\prime=0}(t^\prime+1)^{-1} \leq c^2T^{-1}\cdot(1+\log T) \,.
\end{split}
\end{equation}
Hence,
\begin{equation}
\begin{split}
    &\mathbb{E}[F_{\mathcal{S}}(x_T)-F_{\mathcal{S}}(x^*)] = \widetilde{\mathcal{O}}(T^{-\frac{\mu\delta c}{2}} + T^{-1} + (1/\sqrt{1-\delta}-1)^{-2}) \,.
\end{split}
\end{equation}

\section{Proof of Generalization of SGD-(IA) (Theorem \ref{sgd-ia generalization})}\label{appendix:ia}
For consistency, let $\{y_t\}$ be the SGD solution path, i.e.,
\begin{equation}
    y_{t+1}=y_t-\eta_t\nabla f(y_t;\xi_t)=y_t-\eta_t g_t \,.
\end{equation}
Let $\{x_t\}$ be the SGD-IA solution path with $x_0=y_0$, where IA denotes momentum iterative averaging, i.e.,
\begin{equation}
    x_{t+1}=(1-\delta)x_t+\delta y_{t+1} = x_t + \delta(y_{t+1}-x_t) \text{ and } x_0=y_0 \,,
\end{equation}
where $0<1-\delta<1$ is the momentum constant. Then,
\begin{equation}
    x_t=(1-\delta)^tx_0 + \sum^{t}_{t^\prime=1}\delta(1-\delta)^{t-t^\prime}y_{t^\prime} = (1-\delta)^ty_0 + \sum^{t}_{t^\prime=1}\delta(1-\delta)^{t-t^\prime}y_{t^\prime} \,.
\end{equation}
Let $K=1$ and $\lambda=1$, then DEF is identical to SGD. Following Lemma \ref{lemma:ia def}, SGD-IA is a special case of DEF-A with $\mathcal{C}(\Delta)=\delta \Delta$. Note that this compressor does not compress the message volume.
\begin{lemma} \label{lemma:ia def}
Let $g_t=\nabla f(y_t;\xi_t)$, $x_0=y_0$, and $\mathcal{C}(-\Delta)=\mathcal{C}(\Delta)$. If $y_{t+1}=y_t-\eta_tg_t$ and $x_{t+1}=x_t+\mathcal{C}(y_{t+1}-x_t)$, then the update rule of $x_t$ is identical to DEF-A when $K=1$, $\lambda=1$ with compressor $\mathcal{C}$, i.e.,
\begin{equation}
\begin{split}
    x_{t+1} &= x_t - \mathcal{C}(\eta_tg_t+e_t) \,, \\
    e_{t+1} &= \eta_t g_t+e_t -\mathcal{C}(\eta_tg_t+e_t) ,\, e_0=0 \\
    y_{t+1} &= y_t -\eta_t g_t \,.
\end{split}
\end{equation}
\end{lemma}
\begin{proof}
We just need to verify $x_{t+1}=x_t+\mathcal{C}(y_{t+1}-x_t)$ with the 3 equations above. We have
\begin{equation}
    x_{t+1} = x_0 - \sum^{t}_{t^\prime=0}\mathcal{C}(\eta_{t^\prime}g_{t^\prime}+e_{t^\prime}),\, y_{t+1} = y_0-\sum^{t}_{t^\prime=0}\eta_tg_{t^\prime} \,.
\end{equation}
Then
\begin{equation}
\begin{split}
    x_{t+1} - e_{t+1} &= x_0 - e_{t+1} - \sum^{t}_{t^\prime=0}\mathcal{C}(\eta_{t^\prime}g_{t^\prime}+e_{t^\prime}) \\
    &= x_0 - e_{t+1} - \mathcal{C}(\eta_{t}g_{t}+e_{t}) - \sum^{t-1}_{t^\prime=0} \mathcal{C}(\eta_{t^\prime}g_{t^\prime}+e_{t^\prime}) \\
    &= x_0 - \eta_tg_t - e_t - \sum^{t-1}_{t^\prime=0} \mathcal{C}(\eta_{t^\prime}g_{t^\prime}+e_{t^\prime}) \\
    &=\cdots=x_0-\sum^{t}_{t^\prime=0}\eta_tg_t = y_{t+1} \,,
\end{split}
\end{equation}
\begin{equation}
\begin{split}
    x_t+\mathcal{C}(y_{t+1}-x_t) &= x_t + \mathcal{C}(x_{t+1}-e_{t+1}-x_t) \\
    &= x_t + \mathcal{C}(-\mathcal{C}(\eta_tg_t+e_t)-e_{t+1}) \\
    &= x_t + \mathcal{C}(-\eta_tg_t-e_t)=x_{t+1} \,,
\end{split}
\end{equation}
which completes the proof.
\end{proof}

\subsection{Generalization Error of SGD}
In this section, we need Assumptions \ref{bounded variance}, \ref{bounded second moment}, \ref{lipschitz gradient}, and $\eta_t\leq\frac{c}{t+1}$. Following Sec.~\ref{sec:def gen} with $K=1$ and $\lambda=1$, we have
\begin{equation}
    \mathbb{E}[f(y_T;\xi)-f(\widetilde{y}_T;\xi)] = \mathcal{O}(T^{(1-\frac{1}{N})Lc/((1-\frac{1}{N})Lc+1)}) \,.
\end{equation}

\subsection{Generalization Error of SGD-IA}
In this section, we need Assumptions \ref{bounded variance}, \ref{bounded second moment}, \ref{lipschitz gradient}, and $\eta_t\leq\frac{c}{t+1}$. Following Sec.~\ref{sec:defa gen} with $K=1$, $\lambda=1$ and the compressor replaced with $\mathcal{C}(\Delta)=\delta\Delta$ which satisfies Assumptions \ref{approximate compressor} and \ref{allreduce compressor}, we have $\delta^{\frac{1}{2}}\rightarrow \delta$ in Eq.~(\ref{eq:defa gen bound 1}). All the other procedures are the same, thus
\begin{equation}
    \mathbb{E}[f(x_T;\xi)-f(\widetilde{x}_t;\xi)] = \mathcal{O}(T^{(1-\frac{1}{N})\delta Lc/((1-\frac{1}{N})\delta Lc+1)}) \,.
\end{equation}

\subsection{Optimization Error of SGD}
In this section, we need Assumptions \ref{bounded variance}, \ref{bounded second moment}, \ref{lipschitz gradient}, and $\eta_t=\frac{c}{t+1}\leq \frac{1}{4L}$. Following Sec.~\ref{sec:def opt} with $K=1$ and $\lambda=1$, we have
\begin{equation}
    \mathbb{E}[F_{\mathcal{S}}(y_T)-F_{\mathcal{S}}(y^*)]=\widetilde{\mathcal{O}}(T^{-\frac{\mu c}{2}}+T^{-1}) \,.
\end{equation}

\subsection{Optimization Error of SGD-IA}
In this section, we need Assumptions \ref{bounded variance}, \ref{bounded second moment}, \ref{lipschitz gradient}, and $\eta_t=\frac{c}{t+1}\leq \frac{1}{8\delta L}$. Following Sec.~\ref{sec:defa opt} with $K=1$, $\lambda=1$ and the compressor replaced with $\mathcal{C}(\Delta)=\delta\Delta$ which satisfies Assumptions \ref{approximate compressor} and \ref{allreduce compressor}, we have the same bound as Eq.~(\ref{eq:defa opt 1}), but $\tcircle{2}=\delta^2\mathbb{E}\|\eta_tg_t+e_t\|^2_2$ in Eq.~(\ref{eq:defa opt 2}), which leads to the need for $\eta_t\leq\frac{1}{8\delta L}$. All the other procedures are the same, therefore
\begin{equation}
\begin{split}
    &\mathbb{E}[F_{\mathcal{S}}(x_T)-F_{\mathcal{S}}(x^*)] = \widetilde{\mathcal{O}}(T^{-\frac{\mu\delta c}{2}} + T^{-1} + (1/\sqrt{1-\delta}-1)^{-2}) \,.
\end{split}
\end{equation}